\documentclass[runningheads]{llncs}
\usepackage{titletoc}
\usepackage{graphicx}
\usepackage{booktabs}
\usepackage{multirow}

\usepackage[accsupp]{axessibility}
\usepackage{hyperref}
\usepackage[numbers,sort&compress]{natbib}
\usepackage{array}

\usepackage{xspace}
\newcommand{\shortname}{SpectralSplats\xspace}

\usepackage{orcidlink}
\usepackage{wrapfig}
\usepackage{bm}
\usepackage{amsmath}
\usepackage{subcaption}
\usepackage{geometry}
\begin{document}
\addtocontents{toc}{\protect\setcounter{tocdepth}{-10}}

\title{\shortname: Robust Differentiable Tracking via Spectral Moment Supervision} 

\titlerunning{SpectralSplats}

\author{Avigail Cohen Rimon\inst{1}\orcidlink{0009-0000-7080-6091} \and
Amir Mann\inst{1}\orcidlink{0009-0006-9821-2758} \and
Mirela Ben-Chen\inst{1}\orcidlink{0000-0002-1732-2327} \and
Or Litany\inst{1,2}\orcidlink{0000-0001-6700-7379}}

\authorrunning{SpectralSplats}

\institute{Technion - Israel Institute of Technology\and
NVIDIA
}

\maketitle
\begin{figure*}
    \small
    \centering    
    \includegraphics[width=0.99\textwidth,trim={0 0.6cm 0 0},clip]{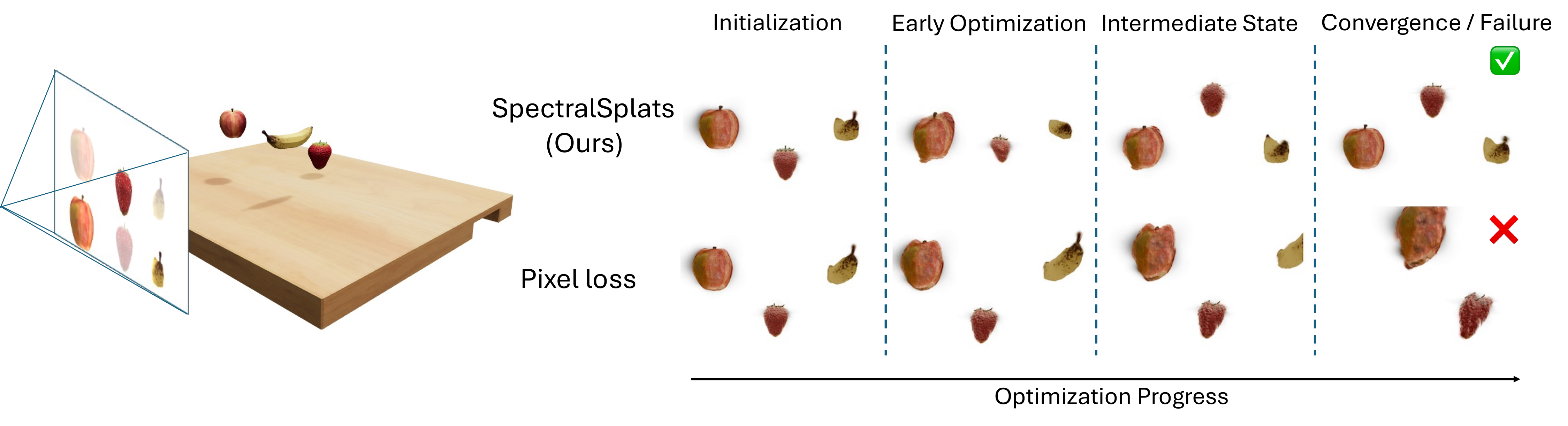}
    \vspace{-10pt}
    \caption{\textbf{\shortname enables robust tracking from zero-overlap initializations.} \textit{Left:} A 3DGS asset is initialized (see transparent overlay) far from some target
pose image (solid image), resulting in strictly zero spatial overlap in the rendered camera view. \textit{Right:} We compare the optimization progression. Standard photometric tracking (Pixel loss) implicitly requires spatial overlap; without it, directional gradients vanish, causing the optimizer to strand the asset and eventually collapse into spurious local minima. \shortname (Ours) shifts supervision to the frequency domain via \textit{Spectral Moments}. This establishes a global basin of attraction, allowing the Gaussians to smoothly flow across the image domain and successfully recover the extreme displacement.}
    \label{fig:teaser}
\end{figure*}
\vspace{-30pt}
\begin{abstract}
3D Gaussian Splatting (3DGS) enables real-time, photorealistic novel view synthesis, making it a highly attractive representation for model-based video tracking. However, leveraging the differentiability of the 3DGS renderer ``in the wild'' remains notoriously fragile. A fundamental bottleneck lies in the compact, local support of the Gaussian primitives. Standard photometric objectives implicitly rely on spatial overlap; if severe camera misalignment places the rendered object outside the target's local footprint, gradients strictly vanish, leaving the optimizer stranded. We introduce SpectralSplats, a robust tracking framework that resolves this "vanishing gradient" problem by shifting the optimization objective from the spatial to the frequency domain. By supervising the rendered image via a set of global complex sinusoidal features (Spectral Moments), we construct a global basin of attraction, ensuring that a valid, directional gradient toward the target exists across the entire image domain, even when pixel overlap is completely nonexistent. To harness this global basin without introducing periodic local minima associated with high frequencies, we derive a principled Frequency Annealing schedule from first principles, gracefully transitioning the optimizer from global convexity to precise spatial alignment. We demonstrate that SpectralSplats acts as a seamless, drop-in replacement for spatial losses across diverse deformation parameterizations (from MLPs to sparse control points), successfully recovering complex deformations even from severely misaligned initializations where standard appearance-based tracking catastrophically fails.
\end{abstract}

\section{Introduction}
\label{sec:intro}

The recent advent of 3D Gaussian Splatting (3DGS) \citep{kerbl20233d} has fundamentally disrupted the landscape of 3D reconstruction. By representing scenes as a collection of anisotropic 3D Gaussians, 3DGS achieves real-time rendering speeds and photorealistic quality. On top of being exceptionally capable at \textit{static} Novel View Synthesis (NVS), its differentiable rendering property enables a critical application: the ability to take a reconstructed static asset and "enact" it by fitting it to a target video~\cite{luiten2024dynamic,lei2024gart,bekor2025gaussian}. 

This task of model-based video tracking -- estimating continuous geometric motion parameters to match a target observation -- is foundational for applications like driving digital avatars, markerless motion capture, and editable dynamic scenes. Yet, estimating these continuous geometric displacements purely from visual observation remains an open and highly fragile challenge.

The core difficulty lies in the optimization landscape of \textit{Analysis-by-Synthesis}. In a typical model-based tracking pipeline, we seek the motion parameters $\theta$ that minimize the photometric error between the rendered model and the observed target. This optimization relies on the differentiability of the renderer to backpropagate gradients from pixel errors to motion parameters. Crucially, this mechanism relies on \textit{local} spatial overlap: for a primitive to receive gradient updates towards its corresponding visual structure in the target image, its rendered footprint must already intersect with that structure's location. Since Gaussian splats are local primitives with compact support, if the estimated motion parameters are sufficiently far from the target (e.g., due to a coarse initialization or noisy pose priors), the rendered Gaussians do not overlap with their intended target pixels. As illustrated in Fig.~\ref{fig:teaser}, without this directional signal, the gradient component corresponding to the true target vanishes $(\nabla_{\Theta}\mathcal{L}_{target}\rightarrow0)$, and the optimizer is actively steered towards arbitrary distractors or irrelevant local minima rather than the correct solution. 

Fig.~\ref{fig:convergence_analysis} dissects this "vanishing gradient" pathology in 1D. Under large spatial displacements, the standard spatial $L_2$ landscape lacks a global basin leading to the correct state, causing the tracker to fail catastrophically. 

A standard workaround for this "basin of attraction" problem in dynamic 3D reconstruction is to rely on manual alignment or controlled setups to guarantee sufficient spatial overlap from the very first frame. Recent approaches like \cite{bekor2025gaussian} found it useful to replace the standard $L_2$ loss with deep feature distances such as LPIPS. While the hierarchical receptive fields of these networks moderately widen the basin of attraction compared to raw pixel errors, they still fundamentally rely on localized spatial overlap. Under severe camera misalignments or rapid motion where the rendered asset and the target are disjoint, the gradients from these deep features still vanish. Alternatively, approaches relying on \textit{category-specific priors}~\cite{loper2023smpl,zuffi20173d} bypass the global search problem by leveraging off-the-shelf pose estimators to provide a strong initial alignment, ensuring sufficient spatial overlap before appearance-based optimization even begins. While this reduces the photometric tracking to a simple ``last-mile'' refinement, 
it achieves robustness only by sacrificing generality, rendering them unsuitable for tracking arbitrary, ``in-the-wild'' objects. Consequently, there remains a critical need for a purely optimization-based tracking objective that is both global (capable of handling large, disjoint displacements) and class-agnostic.

To bypass this initialization dependency, we introduce \textit{\shortname}, a robust tracking framework that solves the vanishing gradient problem through Spectral Moment supervision. Our key insight is to shift the optimization objective from the spatial domain to the frequency domain. Unlike pixels or rendered splats, which are local, sinusoidal basis functions are global. By projecting the rendered image onto a set of complex Fourier features, we compute a ``spectral signature'' of the current pose. A spatial displacement of the object corresponds to a phase shift in these frequencies, providing a strong, non-zero gradient signal even when the object and its target are spatially disjoint.

To successfully harness this global basin, we employ a rigorous \textbf{coarse-to-fine Frequency Annealing strategy}. We establish that while low-frequency moments provide the long-range attraction necessary for global tracking, they lack fine grained precision. By dynamically adjusting the active frequency bandwidth—systematically transitioning from coarse boundaries to precise structural alignments—we guide the underlying tracker into an  accurate final pose.
Our spectral loss serves as a general-purpose objective function that is agnostic to the underlying deformation model. We demonstrate its efficacy on two prevalent non-rigid parameterizations: sparse control points driven continuously by neural MLPs \cite{yang2024deformable}, and control points optimized directly via explicit displacements \cite{huang2024sc}.
By integrating our global supervision into these distinct architectures, we show that it can guide the underlying tracker from extreme initial displacements -- which cause standard photometric losses to fail -- towards a highly accurate final pose, without requiring modifications to the deformation models themselves.

Our contributions are:
\begin{itemize}
    \renewcommand\labelitemi{$\bullet$}
    \item \textbf{Spectral Moment Loss:} A novel, global objective function for 3DGS that provides non-vanishing directional gradients, effectively eliminating the ``vanishing gradient'' problem inherent to localized photometric losses under large spatial misalignments.
    
    \item \textbf{Principled Frequency Annealing:} A systematic optimization schedule derived from a first-principles analysis of phase wrapping. By progressively expanding the active frequency bandwidth from coarse to fine, we effectively smooth the high-frequency ambiguities of the spatial loss landscape. This significantly broadens the basin of attraction, bridging large spatial misalignments before refining high-frequency structural details.

    \item \textbf{Initialization-Robust Tracking:} We demonstrate the versatility of our global formulation across both synthetic and real-world datasets. By seamlessly integrating our spectral loss with diverse deformation representations (MLPs and sparse control points) and standard local objectives ($L_2$ and LPIPS), we consistently improve tracking stability. Our method successfully recovers complex deformations and survives severe camera misalignments, where standard appearance-based objectives fail.
\end{itemize}
\section{Related Work}
\label{sec:related}
The development of SpectralSplats intersects with two primary research trajectories: the parameterization of Dynamic 3D Scene Reconstruction, and the shaping of Frequency-Guided Optimization Landscapes.
  
\subsection{Dynamic and Deformable 3D Gaussian Splatting}

Following the seminal work on static 3DGS~\citep{kerbl20233d}, splat-based representations were rapidly extended to \emph{dynamic} scenes~\cite{yang2024deformable,luiten2024dynamic,duisterhof2023deformgs,seidenschwarz2025dynomo,kratimenos2024dynmf,li2024spacetime,sun20243dgstream,Wu_2024_CVPR,yang2023real,yan2024street, zhou2024drivinggaussian, zhou2024hugs, chen2024omnire}. The core challenge is to model the temporal evolution of Gaussian parameters while preserving temporal coherence. A dominant paradigm is \emph{canonicalization}, which pairs a static canonical set of Gaussians with a time-varying deformation model. Such systems are typically trained either end-to-end from video \citep{zhou2024drivinggaussian, zhou2024hugs,chen2024omnire} or via a two-stage pipeline that first initializes a canonical representation and then tracks per-frame deformations \citep{Wu_2024_CVPR,duisterhof2023deformgs}. Our setting aligns with the latter: we focus on deformation-based matching across frames, assuming a reliable initialization of the canonical scene.

Tracking dynamic scenes is inherently under-constrained and prone to geometric artifacts. To make tracking tractable and enforce temporal coherence, prior work commonly injects structural priors into the deformation model. Coordinate-based MLPs are frequently used to learn continuous displacement fields, prioritizing smoothness and coherence \citep{yang2024deformable, li2024spacetime, sun20243dgstream}. To accelerate training and inference speeds, approaches utilize structured grid encodings \cite{Wu_2024_CVPR, cao2023hexplane,fridovich2023k}. To further regularize these fields, recent methods have moved toward explicit geometric constraints like sparse control points~\cite{huang2024sc,bekor2025gaussian}, while DynMF~\cite{kratimenos2024dynmf} utilizes low-dimensional neural motion factorization. 
Recent advancements in online tracking have further pushed the boundaries of this paradigm; \cite{jin20256dope} utilizes incremental 2D Gaussian Splatting~\cite{huang20242d} for efficient online 6-DoF object pose estimation, while FeatureSLAM~\cite{thirgood2026featureslam} integrates foundation model features into the 3DGS rasterization pipeline for real-time semantic tracking.

While these structural design choices improve temporal consistency and rendering quality, they fundamentally assume that gradients from a photometric objective remain informative. Consequently, they do not resolve the optimization failure that occurs when the rendered object is spatially disjoint from its true image location. Our \shortname framework is complementary to these motion models; it provides a global supervisory signal that can guide any of the aforementioned parameterizations toward alignment from poor initializations.

To bypass this global search problem, domain-specific parameterizations heavily restrict the solution space. Human-centric methods such as HUGS \cite{kocabas2024hugs} leverage SMPL \cite{loper2023smpl} to optimize body pose deformations. Similarly, GART \cite{lei2024gart} proposes a canonical articulated template, extending the rigidity of bone-transformations to 3DGS primitives. While these articulated priors yield strong performance when the category assumption holds, they are brittle to initialization errors that place the template outside the local photometric basin. Our \shortname framework is complementary to these motion models; it provides a global supervisory signal that can guide any of the aforementioned parameterizations toward alignment from poor initializations.

\subsection{Frequency Analysis and Annealing in Neural Rendering}
The interplay between spectral analysis and neural optimization has been a focal point of recent research, particularly regarding the "spectral bias" of neural networks. While high-frequency components are essential for capturing fine-grained detail, they often induce a rugged loss landscape, complicating the optimization of geometric parameters.

\noindent\textbf{Frequency for Representation Quality.} To mitigate these instabilities, several works have proposed managing spectral bandwidth to improve reconstruction fidelity. In the implicit domain, SAPE~\cite{Hertz2021SAPE} modulates the frequency of positional encodings spatially, preventing noise-induced minima in smooth regions.
With the shift to explicit Gaussian primitives, similar principles have been applied to regularize structure: FreGS~\cite{Zhang2024FreGS} employs progressive frequency regularization to mitigate densification artifacts, while PGDGS~\cite{huang2025pgdgs} adopts progressive Gaussian densification for sparse-view reconstruction.
Lavi et al.~\cite{lavi2025frequency} structure the scene into hierarchical Laplacian pyramid subbands to decouple low-frequency geometry from high-frequency residuals.
Crucially, these methods leverage frequency decomposition primarily for \textit{level-of-detail} control and static representation quality.

\noindent\textbf{Frequency for Geometric Optimization.} Beyond representation, frequency analysis offers a powerful tool for shaping the optimization landscape. In the context of NeRF~\cite{mildenhall2021nerf}, BARF~\cite{lin2021barf} utilized spectral annealing of positional encoding to widen the basin of attraction for camera registration, while MomentsNeRF~\cite{almughrabi2024momentsnerf} leveraged moment constraints for few-shot supervision. We transpose these insights to the domain of dynamic 3DGS. However, rather than annealing positional encodings, we propose Spectral Moment Supervision directly on the rendered output. This effectively bypasses the vanishing gradient problem inherent in spatial losses, creating a global basin of attraction that guides Gaussians even from zero-overlap initializations. Crucially, to avoid the phase-wrapping traps inherent in high frequencies, we introduce a principled Frequency Annealing schedule. While prior methods motivated linearly scaling frequency schedules heuristically through Neural Tangent Kernel \cite{jacot2018neural} theory or signal bandwidth blurring~\cite{park2021nerfies, lin2021barf} we formally derive our annealing schedule from first principles.

\section{Method}
\label{sec:method}

We present \shortname, a framework for robust dynamic tracking that replaces standard spatial photometric errors with a spectral objective. We first formalize the ``vanishing gradient'' failure mode inherent to 3DGS tracking, establish the spectral-spatial duality of our objective, and then introduce our principled Spectral Moment Supervision and Frequency Annealing schedule.

\subsection{Differentiable Gaussian Tracking and the Vanishing Gradient}

A 3D Gaussian Splatting scene is parameterized by a set of primitives $\mathcal{G} = \{G_i\}$, each defined by a 3D mean $\mu_i$, covariance $\Sigma_i$, opacity $\alpha_i$, and spherical harmonics coefficients $c_i$. The rasterization function $\mathcal{R}$ projects these 3D primitives onto the 2D image plane to produce a rendering $\mathbf{I}_{\text{rend}}$. 

In a tracking context, we assume a static canonical model $\mathcal{G}_{\text{ref}}$ is given. We seek a set of motion parameters $\Theta$ (e.g., representing a rigid transformation $\mathbf{T} \in SE(3)$ or neural deformation weights) that parameterize a deformation function $\mathcal{D}$. This function acts on the canonical model to produce a displaced scene: $\mathcal{G}_{\text{def}} = \mathcal{D}(\mathcal{G}_{\text{ref}}; \Theta)$. The rasterization function $\mathcal{R}$ then projects these deformed 3D primitives onto the 2D image plane to produce the rendering $\mathbf{I}_{\text{rend}}(\mathbf{p}; \Theta) = \mathcal{R}(\mathcal{D}(\mathcal{G}_{\text{ref}}; \Theta))$, which we aim to align with an observed target image $\mathbf{I}_{\text{gt}}$. To formally analyze the optimization landscape, we treat the image domain continuously and define the standard objective as minimizing the photometric difference over all 2D spatial coordinates $\mathbf{p}$:
\begin{equation}
\mathcal{L}_{\text{photo}}(\Theta) = \frac{1}{2} \int | \mathbf{I}_{\text{rend}}(\mathbf{p}; \Theta) - \mathbf{I}_{\text{gt}}(\mathbf{p}) |_2^2 d\mathbf{p}
\end{equation}

\noindent\textbf{The Vanishing Gradient Problem.} 
\begin{figure}[t]
  \centering
  \includegraphics[width=\linewidth]{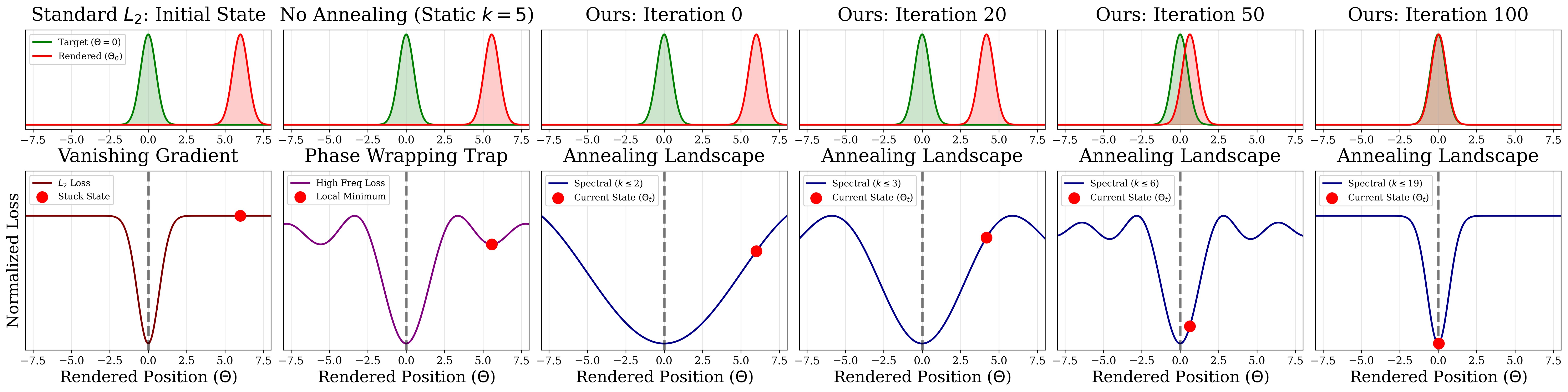}
  \caption{\textbf{Breaking the Locality Trap: A 1D Optimization Analysis.} We simulate the optimization landscape (bottom) for aligning a rendered 1D Gaussian pulse (top, red) to a target (top, green) under a large initial spatial displacement ($\Theta_0 = 6$). \textbf{Standard $\mathbf{L_2}$ (Col 1):} Photometric objectives implicitly rely on spatial overlap; without it, the gradient strictly vanishes, leaving the optimizer stranded. \textbf{No Annealing (Col 2):} Projecting the loss onto a static, high-frequency spectral basis ($k=5$) ensures the gradient no longer vanishes globally, but introduces severe phase-wrapping that traps the optimizer in false local minima. \textbf{Ours (Cols 3-6):} Spectral Moment Supervision with Frequency Annealing. By restricting initial supervision to low frequencies, we construct a globally convex basin of attraction that provides a valid, directional gradient from any initialization. As the spatial error strictly decreases, our principled annealing schedule safely expands the active bandwidth, seamlessly transitioning the landscape to achieve high-frequency spatial precision without phase-wrapping.}
\label{fig:convergence_analysis}  
\end{figure}
Before diving into the formal analysis, the core intuition behind this failure mode is remarkably simple: standard photometric tracking compares pixels locally. Because a Gaussian primitive only influences a compact spatial footprint, it must physically overlap with the target structure to receive a meaningful update. If the initial displacement is large enough that there is strictly zero overlap, moving the Gaussian slightly in any direction does not alter the total image loss. Because a small local translation yields absolutely zero change in the photometric error, the gradient evaluates exactly to zero. The loss is high, but as simulated in Fig.~\ref{fig:convergence_analysis} (Col 1), the local optimization landscape is entirely flat, leaving the optimizer stranded. 

To rigorously derive this ``locality trap'', let us isolate the optimization of a rendered Gaussian and its corresponding true target signal in the image. By expanding the derivative of the squared error for this source-target pair, we can decompose its gradient contribution into two distinct components:

\begin{equation}
\nabla_\Theta \mathcal{L}_{\text{photo}}^{\text{target}} = \underbrace{\int \mathbf{I}_{\text{rend}}(\mathbf{p}; \Theta) \nabla_\Theta \mathbf{I}_{\text{rend}}(\mathbf{p}; \Theta) d\mathbf{p}}_{\text{Self-Term}} - \underbrace{\int \mathbf{I}_{\text{gt}}(\mathbf{p}) \nabla_\Theta \mathbf{I}_{\text{rend}}(\mathbf{p}; \Theta) d\mathbf{p}}_{\text{Target Supervision}}
\end{equation}

This decomposition highlights the fundamental flaw in tracking with highly localized spatial functions. The \textit{Self-Term} can be rewritten via the chain rule as $\frac{1}{2} \nabla_\Theta \int \mathbf{I}_{\text{rend}}^2 d\mathbf{p}$. 
For translations parallel to the image plane, this operation preserves the total footprint mass of the rendered object, making the integral strictly invariant to the motion parameter $\Theta$ and its derivative exactly zero. While depth translations do yield a non-zero derivative due to perspective projection, in the absence of target overlap, this gradient merely acts to minimize the rendering footprint, driving the object to shrink by moving far away from the camera. In neither case does the self-term provide a directional signal toward the true target.
The tracking signal therefore relies entirely on the \textit{Target Supervision} cross-term. However, if $\Theta$ positions the rendered Gaussian such that it is spatially disjoint from its true target location in $\mathbf{I}_{\text{gt}}$, the product of $\mathbf{I}_{\text{gt}}(\mathbf{p})$ and the spatial boundaries of the rendered object $\nabla_\Theta \mathbf{I}_{\text{rend}}(\mathbf{p}; \Theta)$ is zero everywhere. Consequently, the gradient contribution pulling the object to its true destination vanishes completely. This mathematical trap is further enforced by the 3DGS architecture itself: to maintain real-time performance, the rasterizer splits the screen into $16 \times 16$ tiles and culls primitives using a 99\% confidence interval \cite{kerbl20233d}, forcefully zeroing out gradients for targets outside the immediate tile vicinity.

Crucially, this vanishing gradient means that even though the photometric error is large, the loss cannot decrease because the local gradient landscape is flat. Worse, when viewing the entire loss $\mathcal{L}_{\text{photo}}$ over a complete scene, the overall gradient $\nabla_\Theta \mathcal{L}_{\text{photo}}$ does not evaluate to zero. The misaligned Gaussian inevitably overlaps with incorrect content in $\mathbf{I}_{\text{gt}}$ (e.g., background clutter). Because the true gradient has vanished, the optimizer receives only corrupted gradients driven by this spurious overlap. Rather than pulling the Gaussian toward its target, these gradients anchor the object to the background.

\subsection{Image Moments and Spectral Duality}\label{subsec:moments_and_duality}

To resolve the strict locality of the spatial loss, we shift our objective from direct pixel-to-pixel comparisons to the alignment of \textit{image moments}. Intuitively, computing a moment is equivalent to multiplying the image by an auxiliary static field $F(\mathbf{p})$ and integrating the result. If we choose a field that varies continuously across the entire spatial domain -- such as a sinusoidal wave or a polynomial function -- this projection acts as a global coordinate system.

This global integration breaks the locality trap. Let us define a simple moment-matching objective between the rendered image and the target: 
\begin{equation}
\mathcal{L}_{\text{moment}}(\Theta) = \frac{1}{2} \big( M_{\text{rend}}(\Theta) - M_{\text{gt}} \big)^2,
\end{equation}
where $M = \int \mathbf{I}(\mathbf{p}) F(\mathbf{p}) d\mathbf{p}$ denotes the projection of an image $\mathbf{I}$ onto the field.

The gradient of this objective with respect to the motion parameters is: 
\begin{equation}
\nabla_\Theta \mathcal{L}_{\text{moment}}^{\text{target}} = \big( M_{\text{rend}}(\Theta) - M_{\text{gt}} \big) \nabla_\Theta M_{\text{rend}}(\Theta).
\end{equation}
Unlike the spatial cross-term that vanished, this gradient consists of two reliably non-zero components. First, provided the global field $F(\mathbf{p})$ does not repeat values across the spatial domain, the scalar projections of the disjoint rendered and target objects will differ, 
ensuring a valid error magnitude: $\big( M_{\text{rend}}(\Theta) - M_{\text{gt}} \big) \neq 0$. (As we will discuss next, guaranteeing this non-repeating property is a central challenge when employing periodic spectral bases). Second, assuming simple translation, the directional vector -- the gradient of the rendered moment itself -- evaluates to:
\begin{equation}
\nabla_\Theta M_{\text{rend}}(\Theta) = \int \nabla_\Theta \mathbf{I}_{\text{rend}}(\mathbf{p}; \Theta) F(\mathbf{p}) d\mathbf{p} = \int \mathbf{I}_{\text{rend}}(\mathbf{p}; \Theta) \nabla_{\mathbf{p}} F(\mathbf{p}) d\mathbf{p},
\end{equation}
where the final equality follows by first applying the chain rule for spatial translation ($\nabla_\Theta \mathbf{I}_{\text{rend}} = -\nabla_{\mathbf{p}} \mathbf{I}_{\text{rend}}$) and subsequently performing integration by parts. By ensuring the spatial derivative of the field $\nabla_{\mathbf{p}} F(\mathbf{p})$ is non-zero in the region of interest, this integral provides a valid directional signal. Therefore, even if the rendered object and the target are completely disjoint, the optimizer ``feels'' the slope of the field at the object's current location. The scalar difference provides the magnitude of the pull, while the field gradient dictates the direction, enabling robust registration without explicit feature correspondences. While various global kernels exist (e.g., the standard geometric and orthogonal polynomial moments utilized in classic correspondence-free shape alignment \cite{domokos2011nonlinear}), we propose using \textit{Spectral Moments} governed by complex sinusoidal functions, as they provide geometrically meaningful phase shifts under translation. We define a spectral moment for a discrete 2D spatial frequency vector $\mathbf{\omega}_{k_x, k_y}$ (where $k_x$ and $k_y$ are the horizontal and vertical frequency indices) as:

\begin{equation}
\mathcal{M}(k_x, k_y; \mathbf{I}) = \sum_{\mathbf{p}} \mathbf{I}(\mathbf{p}) \cdot \exp(-j \mathbf{\omega}_{k_x, k_y}^T \mathbf{p}).
\end{equation}

Unlike the standard spatial $L_2$ loss, this operation pointwise multiplies the image by a complex sinusoid and integrates it over the entire domain. \footnote{While a naive evaluation over a dense frequency grid is computationally prohibitive, this formulation natively supports highly efficient computation via 2D FFT.} 

\noindent{\textbf{Spectral duality.}} An appealing property of choosing this specific spectral basis is the direct mathematical link it provides back to our original spatial objective. By Parseval's theorem, the sum of squared errors evaluated across a \textit{complete} orthogonal frequency basis is strictly equivalent to the spatial $L_2$ loss. However, this equivalence presents a fundamental paradox: if we were to optimize the full spectral basis simultaneously, the objective would perfectly reconstruct the spatial loss landscape, thereby inheriting the exact same vanishing gradient and local minima traps we set out to avoid. As demonstrated in Fig.~\ref{fig:convergence_analysis} (Col 2), high-frequency components introduce severe phase-wrapping that fragments the global basin of attraction, trapping the optimizer in false local minima. Therefore, to harness the non-vanishing global gradients of the spectral domain while ultimately achieving the strict equivalence and precision of the spatial loss, we cannot use the full basis statically; we must dynamically control the active frequency bandwidth during optimization.

\subsection{Deriving the Frequency Annealing Schedule}
To navigate this trade-off between global convergence and spatial precision, we introduce a coarse-to-fine Frequency Annealing schedule. While isolated low frequencies create a global basin of attraction, they inherently lack precision. Because the spatial gradient of a spectral loss scales with the frequency magnitude ($\nabla \mathcal{L} \propto \omega \sin(\omega d)$), the gradient signal of low-frequency moments diminishes as the spatial error $d$ approaches zero. High frequencies are strictly required for fine-grained alignment, but as established, activating them prematurely induces phase-wrapping that traps the optimizer (Fig.~\ref{fig:convergence_analysis}(Col 2)). To achieve global convergence, we must systematically transition from coarse to fine frequencies. As shown in Fig.~\ref{fig:convergence_analysis}(Cols 3-6), this principled progression seamlessly transforms the optimization landscape: it leverages a globally convex basin to rescue the stranded Gaussians from their initial zero-overlap state, and progressively sharpens into a high-precision spatial target without introducing false minima. 

We formalize this Frequency Annealing schedule from first principles. For a spatial misalignment vector $\mathbf{d}_t$ at optimization step $t$, the spectral loss at frequency $\mathbf{\omega}$ is convex only if the induced phase shift does not wrap, i.e., $|\mathbf{\omega}^T \mathbf{d}_t| < \pi$ (see  Supp. Mat. for a detailed derivation). This defines a dynamic stability condition: the maximum active frequency magnitude $||\mathbf{\omega}_{\text{max}}(t)||$ must be inversely proportional to the magnitude of the spatial error $||\mathbf{d}_t||$. When this condition is met and $\mathbf{\omega}^T \mathbf{d}_t$ is small, the spectral loss landscape $E(\mathbf{d}) = 2 - 2\cos(\mathbf{\omega}^T \mathbf{d})$ is well-approximated by a Taylor expansion as a quadratic bowl, $E(\mathbf{d}) \approx (\mathbf{\omega}^T \mathbf{d})^2$. In this strongly convex regime, the gradient is directly proportional to the spatial displacement ($\nabla E \propto \mathbf{d}$). That is, gradient descent naturally takes update steps that scale with the remaining distance to the target. This guarantees the spatial estimation error decays exponentially: $||\mathbf{d}_t|| \leq ||\mathbf{d}_0|| \gamma^t$ for a convergence factor $\gamma \in (0,1)$. To maintain the phase-wrapping constraint $||\mathbf{\omega}_{\text{max}}(t)|| \propto 1/||\mathbf{d}_t||$, the active frequency magnitude must therefore expand exponentially as $\gamma^{-t}$. Because standard spectral grids organize frequencies logarithmically, such that $||\mathbf{\omega}_k|| \propto 2^k$, an exponential growth in frequency magnitude necessitates a strictly \textit{linear} expansion of the active frequency index $k(t)$ over time.

Crucially, this derivation provides a rigorous, first-principles optimization foundation for the empirically successful annealing schedules introduced in Nerfies \cite{park2021nerfies} and later utilized in BARF \cite{lin2021barf}. While prior works motivated a linearly scaling frequency index heuristically -- through the lens of Neural Tangent Kernel (NTK) theory \cite{park2021nerfies} or signal bandwidth blurring \cite{lin2021barf} -- our dynamic phase-wrapping analysis shows exactly \textit{why} it works: a linearly scaling index on a logarithmic grid represents the theoretical upper bound for safe frequency expansion during spatial alignment.

To implement this expansion without injecting discontinuous shocks into the loss landscape, we adopt the smooth cosine weighting function from these works, applying a time-dependent weight $w_k(t)$ to each spectral moment:
\begin{equation}
w_k(t) = \frac{1 - \cos(\pi \cdot \text{clamp}(\alpha(t) - k, 0, 1))}{2}
\end{equation}
where $\alpha(t)$ scales linearly from $0$ to $K$ over the optimization iterations to govern the active bandwidth, and $k \in \{0, \dots, K-1\}$ is the index of the specific frequency band. This formulation allows each successively higher frequency to gracefully fade into the objective and remain active once its transition window is complete.

\begin{wrapfigure}{r}{0.4\textwidth}
    \centering
    \includegraphics[width=0.9\linewidth]{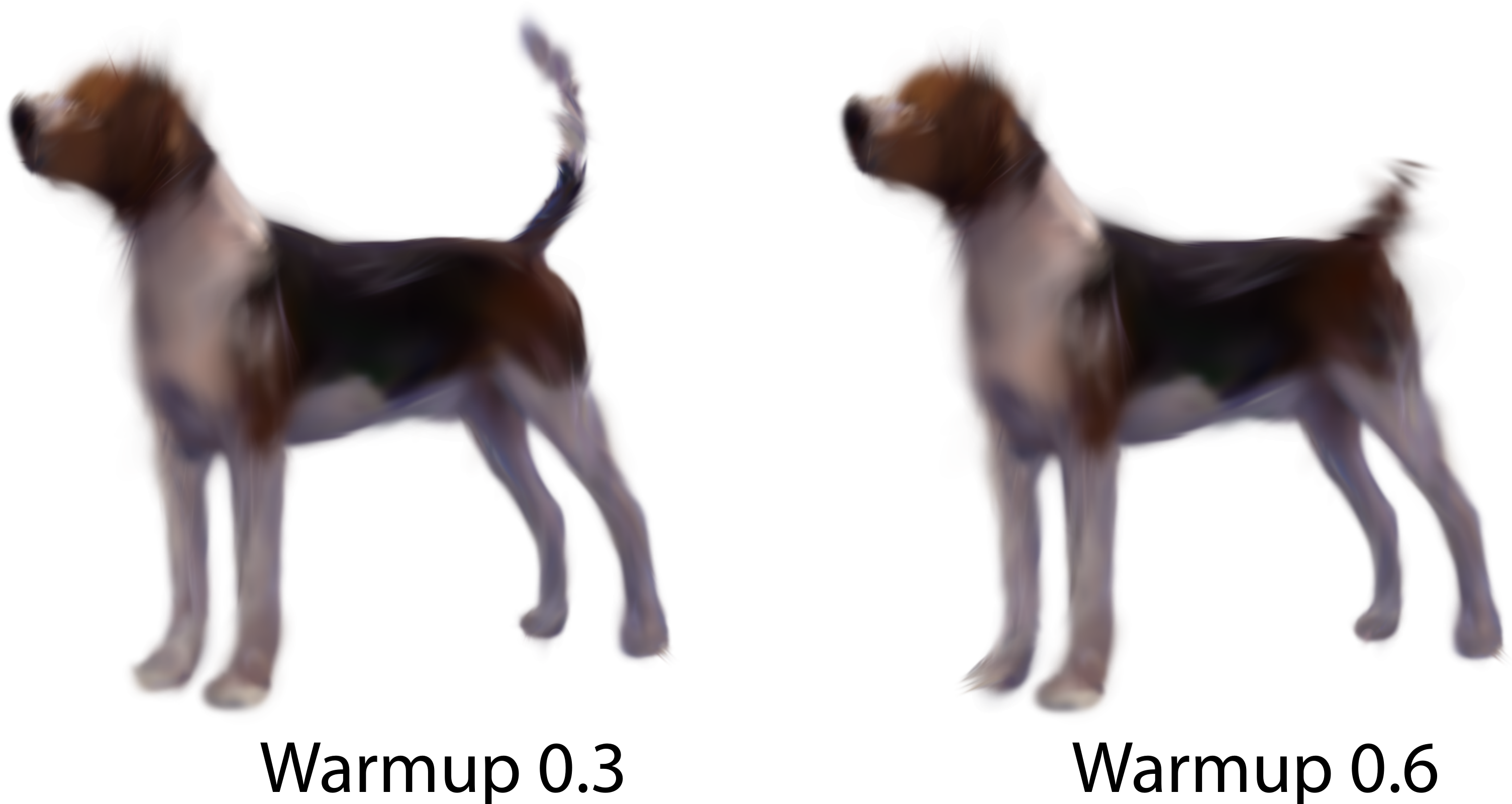}
    \caption{A long low-frequency "warm-up" phase (right) leads to loss of high frequency details (tail), compared to a shorter "warm-up" phase (left).}
    \label{fig:GART_warmup_inset}
\end{wrapfigure}
\noindent\textbf{Conservative Frequency Expansion.} While our derivation establishes that $\alpha(t)$ can safely scale linearly over a logarithmic grid (yielding exponential growth of $\mathbf{\omega}$) under ideal linear convergence, real-world tracking scenarios are rarely ideal.
In practice, background clutter, occlusions, and complex deformations often disrupt ideal exponential error decay.
To account for these unpredictable optimization dynamics, we implement a two-fold conservative scheduling strategy. First, following the empirical practices introduced in BARF~\cite{lin2021barf}, we enforce a strictly low-frequency "warm-up" phase where $\alpha(t)$ is held constant for the initial optimization iterations. This ensures the optimizer has time to exploit the widest global basin and resolve the severe initial spatial misalignments before high-frequency complexities are introduced. Note that if the warm-up period is too long, the high frequency detail may not be recovered correctly (Fig.~\ref{fig:GART_warmup_inset}).
Second, once expansion begins, we linearly scale the frequencies themselves, rather than linearly scaling across their logarithmic indices.
Because linear growth is bounded well below exponential growth, this practical relaxation guarantees that we stay beneath the phase-wrapping threshold ($|\mathbf{\omega}^T \mathbf{d}_t| < \pi$) throughout the optimization process. This delayed, sub-exponential expansion enhances tracking robustness.
We note that our framework operates on foreground Gaussians only. Foreground masks can be obtained once per scene using standard 3D segmentation tools~\cite{fuji2026artisangs}, enabling foreground optimization and background recombination during rendering.

\section{Experiments}
\begin{figure}[t]
    \centering
    \includegraphics[width=.99\textwidth]{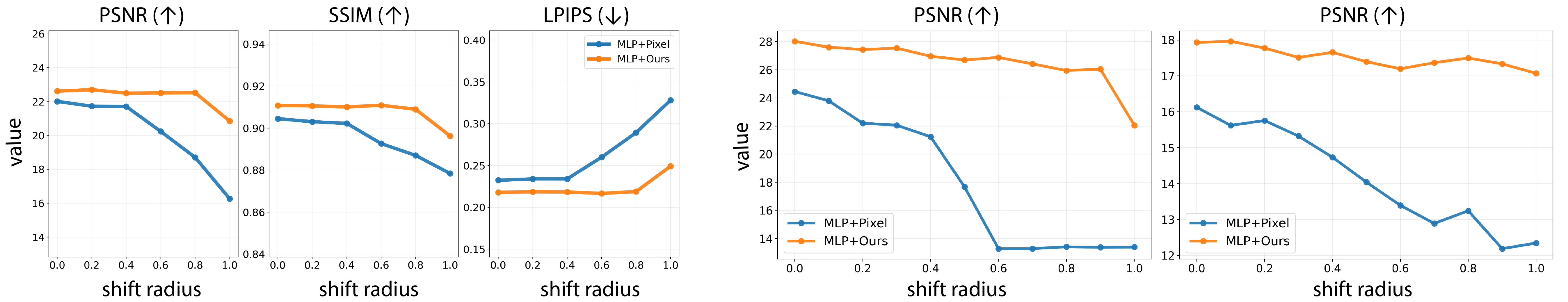}
    \caption{(Left) Effect of spatial shift in GART, showing average PSNR, SSIM, and LPIPS versus shift radius; pixel supervision degrades rapidly, while our method remains stable. (Right) PSNR results on SC4D~\cite{wu2024sc4d, jiang2024consistentd} for training and novel views; pixel loss deteriorates under misalignment, whereas our method maintains stable performance.}
    \label{fig:shift_graphs}
\end{figure}
\noindent{\textbf{Dataset.}}
We evaluate our method on two datasets: 4D Animations generated by SC4D~\cite{wu2024sc4d} using assets from Consistent4D~\cite{jiang2024consistentd}, and the Dog dataset from {GART}~\cite{lei2024gart}.
SC4D provides a controlled setting, where clean and high-quality dynamic 3DGS models are used to render the supervision videos from known views, and provide the ground truth resting 3DGS \cite{kerbl20233d}. As a result, the appearance of the 3DGS initialization is well aligned with the supervision.
To assess performance in a more realistic scenario, we experiment on the GART Dog dataset. Monocular videos, collected from the 2022 National Dog Show and Adobe Stock, are used by GART to reconstruct a unified rest-pose 3DGS model per asset, together the estimated 3DGS and real videos provide our source gaussians and target supervision. This real-world setup includes lighting inconsistencies and unknown camera views, leading to noticeable deviations between the supervision video and the input 3DGS model in pose  and appearance.
As an additional real-world validation, we evaluate our method on a thrown basketball sequence from~\cite{luiten2024dynamic}, demonstrating robustness in a challenging fast-motion scenario.

\noindent{\textbf{Deformation Parameterization.}} Across both datasets, we optimize a deformation model that predicts per-frame displacements of Gaussian \emph{control points}, selected using standard procedures~\cite{huang2024sc}. To evaluate our method across different tracking architectures, we test two variants for moving these control points: (1) \emph{MLP Parameterization}, where a TimeNet~\cite{huang2024sc} network predicts the time-dependent deformation of each control point, and (2) \emph{Direct Morph Field}, where we optimize the positional offsets and rotations of the control points directly.

\noindent{\textbf{Training Objective \& Baselines.}} We implement the Frequency Annealing schedule to resolve initial misalignments. As established in Section \ref{subsec:moments_and_duality}, by Parseval's theorem, optimizing over the full spectral basis is mathematically equivalent to the spatial $L_2$ loss. Therefore, for computational efficiency, once the annealed spectral moments secure local spatial overlap, we naturally transition to standard spatial losses for high-frequency refinement. Specifically, we utilize a pixel loss across both datasets, and LPIPS for the synthetic SC4D dataset. To isolate the contribution of the spectral phase, we compare against baselines relying solely on spatial objectives.
To further evaluate our method, we implemented the global-loss baselines Pyramid of Gaussians (PoG)~\cite{loper2014opendr} and Euclidean Distance Transform (DT)~\cite{liu2019soft}, and compare their performance on both GART (Table~\ref{table_gart_shift}) and the teaser experiment (Supp. Mat.).
To ensure a strictly fair comparison, other loss components and regularization terms follow GSGD~\cite{bekor2025gaussian} and are applied identically to both our method and the baselines (see Supp. Mat.).

\subsection{SC4D Experiment}
\begin{figure}[t]
    \centering
    \includegraphics[width=.9\textwidth]{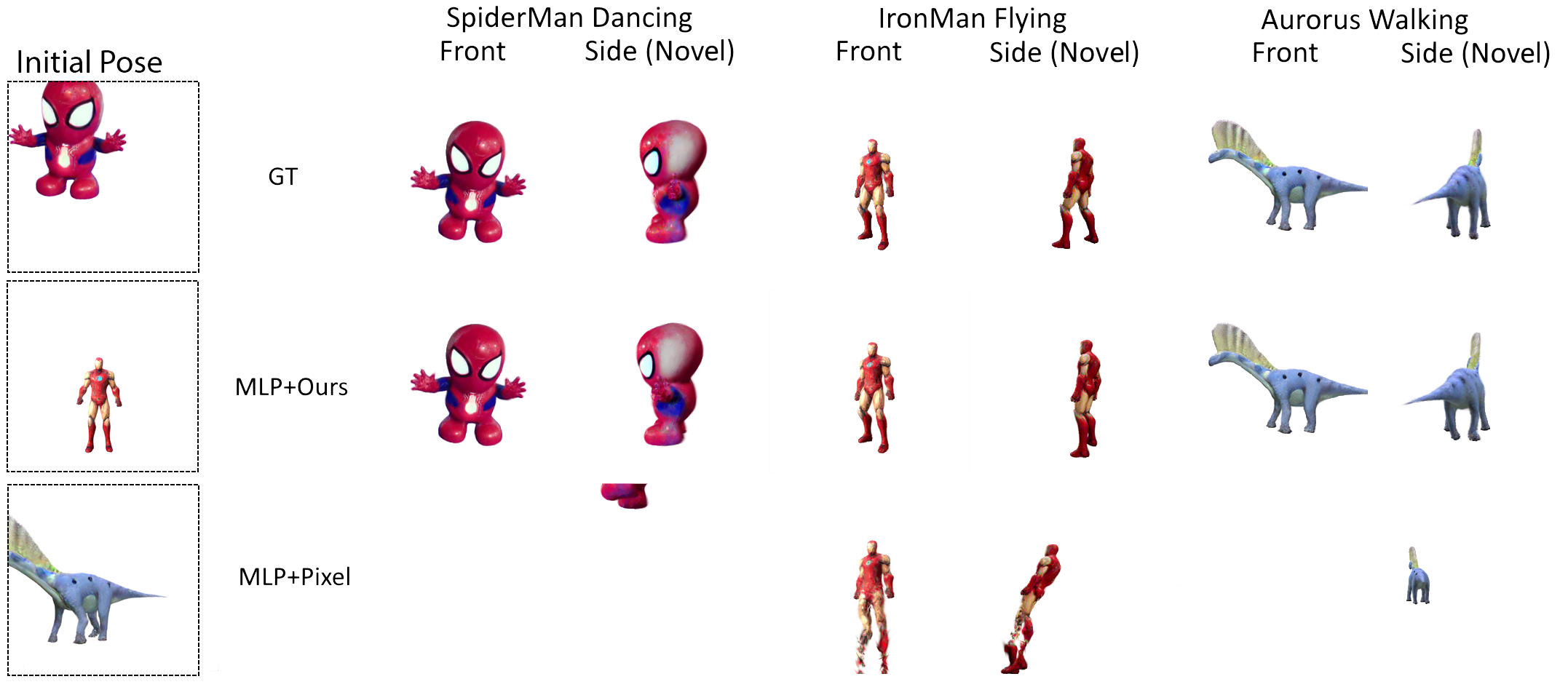}
    \caption{Qualitative comparison on the SC4D data under initial spatial shift (radius = 0.5). For three characters and animations, we show the initial pose, GT, \textit{MLP+Ours} and \textit{MLP+Pixel}, both without LPIPS. While pixel-only optimization fails to recover correct pose and may drift the object outside the frame, our method achieves better alignment and more coherent structure in both training and novel views.}
    \label{fig:SC4D_anymations}
\end{figure}
\begin{table}[t]
\centering
\caption{Metric evaluations on data generated by SC4D \cite{wu2024sc4d} under spatial shift (radius = 0.5). Our method shows large improvement in alignment to the source view across parameterizations and pixel losses, with a decent improvement in novel view quality.}
\label{tab:sc4d_shift0_5_comparison}
\resizebox{\textwidth}{!}{%
\begin{tabular}{l|l|ccc|ccc}
\toprule
& Loss & LPIPS $\downarrow$ & PSNR $\uparrow$ & SSIM $\uparrow$ & NV-LPIPS $\downarrow$ & NV-PSNR $\uparrow$ & NV-SSIM $\uparrow$ \\
\midrule
\multirow{2}{*}{MLP w. LPIPS}
& Pixel & 0.0852 & 23.6051 & 0.9409 & 0.1153 & 18.3244 & 0.9304 \\
& Ours & \textbf{0.0489} & \textbf{27.1453} & \textbf{0.9546} & \textbf{0.0948} & \textbf{19.1977} & \textbf{0.9331} \\
\midrule
\multirow{2}{*}{MLP w/o LPIPS}
& Pixel & 0.1806 & 17.6748 & 0.9108 & 0.2023 & 14.0424 & 0.9107 \\
& Ours & \textbf{0.0516} & \textbf{26.6992} & \textbf{0.9507} & \textbf{0.1331} & \textbf{17.3960} & \textbf{0.9159} \\
\midrule
\multirow{2}{*}{Direct w. LPIPS}
& Pixel & 0.3133 & 11.6619 & 0.8297 & \textbf{0.2443} & 12.3815 & \textbf{0.8727} \\
& Ours & \textbf{0.2000} & \textbf{15.4626} & \textbf{0.8701} & 0.2501 & \textbf{12.7916} & 0.8675 \\
\midrule
\multirow{2}{*}{Direct w/o LPIPS}
& Pixel & 0.2289 & 16.1268 & 0.8562 & 0.2774 & 12.0662 & 0.8491 \\
& Ours & \textbf{0.1868} & \textbf{17.8558} & \textbf{0.8789} & \textbf{0.2640} & \textbf{12.5106} & \textbf{0.8598} \\
\bottomrule
\end{tabular}
}
\end{table}
To simulate spatial misalignment arising from occlusions, drift, or fast motion, we shift the initial 3DGS model in a random direction with increasing offsets, artificially reducing its overlap with the supervision.
We evaluate both pixel-only supervision 
(\textit{MLP+Pixel}) and our spectral scheme (\textit{MLP+Ours}), and report results for both the training and novel views. 
The right panel of Figure~\ref{fig:shift_graphs} shows the mean PSNR as a function of shift radius. As the misalignment increases, the gap between pixel supervision and our method widens. Pixel-based PSNR rapidly decreases, especially under larger shifts, while our method remains considerably more stable. Importantly, this trend holds for both the training and novel views, indicating improved generalization. In the Appendix we provide additional plots evaluating SSIM and LPIPS as a function of shift radius, as well as for supervising with \emph{multiple views}. Crucially, even in the case of multi-view supervision the pixel loss collapses under spatial misalignment, while our \shortname remains robust.

Qualitative results are shown in Figure~\ref{fig:SC4D_anymations} for a representative shift radius of 0.5. We observe that our method remains consistent, whereas pixel-only optimization often fails to recover correct pose and structure, and in some cases even pushes the object outside the frame.
We further report quantitative results in Table~\ref{tab:sc4d_shift0_5_comparison} for this shift radius, evaluated across different deformation parameterizations (MLP and direct morph field). We additionally test replacing the pixel-loss phase with LPIPS supervision. Across almost all configurations, our method consistently improves PSNR, SSIM, and LPIPS in both training and novel views. Overall, these results demonstrate the robustness of our method across parameterizations, spatial loss choices, and evaluation views.

\subsection{GART Experiment}
\begin{figure}[t]
    \centering
    \includegraphics[width=.9\textwidth]{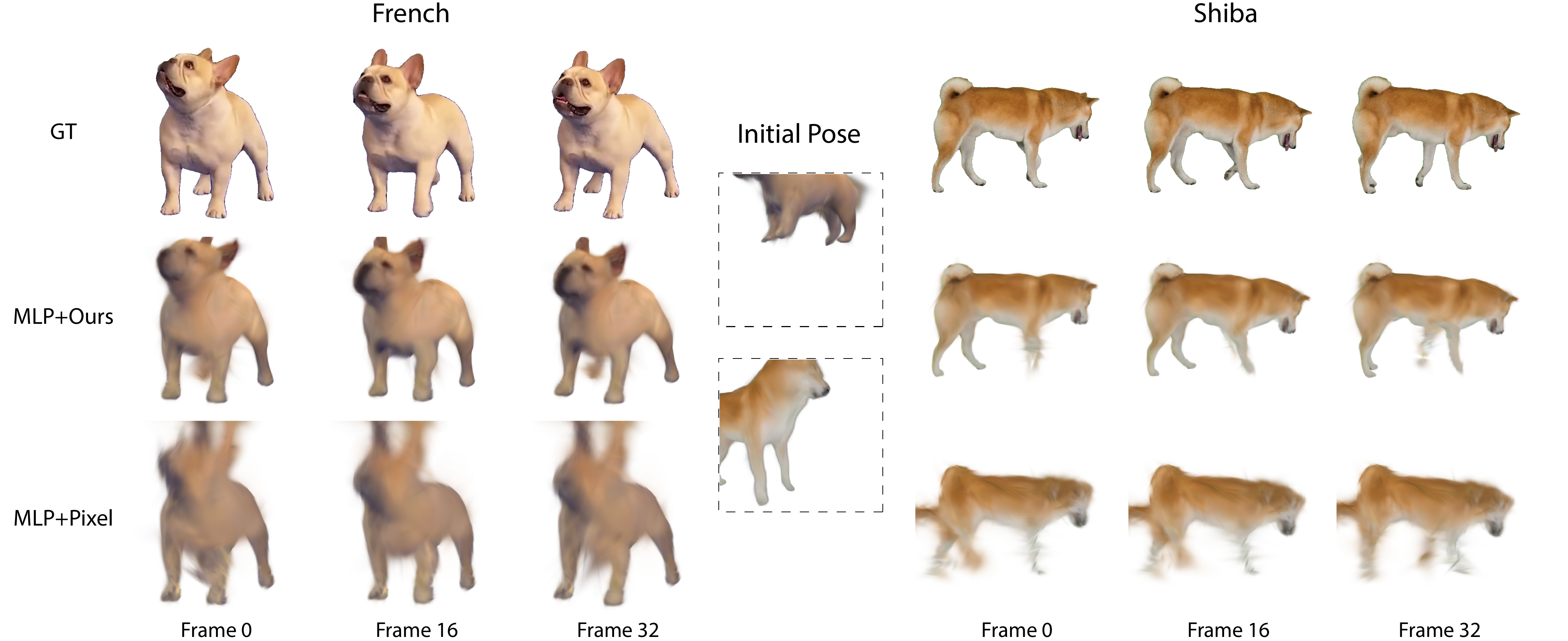}
    \caption{Qualitative comparison under initial spatial misalignment (radius = 0.6) on three frames of two representative dogs from GART. Pixel-only optimization exhibits blur and incorrect alignment, while our method better recovers pose and structure.}
    \label{fig:GART_shift_img}
\end{figure}
We follow a similar spatial misalignment experiment to SC4D, shifting the initial 3DGS model in a random direction with increasing radii. We train using either pixel-only supervision (\textit{MLP+Pixel}) or our spectral scheme (\textit{MLP+Ours}). The left panel of Figure~\ref{fig:shift_graphs} shows the metric means as a function of the shift radius. As the misalignment increases, pixel-only training degrades significantly, while our method remains considerably more stable, highlighting the sensitivity of pixel supervision to poor initialization. In the Appendix we provide per sample plots evaluating PSNR, SSIM and LPIPS as a function of shift radius.

For a representative shift of 0.6, quantitative results are reported in Table~\ref{table_gart_shift} and qualitative comparisons in Figure~\ref{fig:GART_shift_img}. Our \shortname outperforms the pixel baseline on almost all dogs and metrics, improving the mean PSNR (22.05 vs.\ 20.15), SSIM (0.907 vs.\ 0.891), and LPIPS (0.216 vs.\ 0.258). Visual results for \textit{French} and \textit{Shiba} further show better pose recovery and sharper structure with \textit{MLP+Ours}, while \textit{MLP+Pixel} exhibits poor alignment.
Overall, both quantitative and qualitative results demonstrate that our method consistently improves robustness to initial spatial misalignment in realistic 3DGS reconstructions.
\begin{table*}[t]
\centering
\footnotesize
\setlength{\tabcolsep}{3pt}
\caption{GART comparison under a spatial shift radius of 0.6, showing consistent improvements over pixel-only and global-loss baselines (best values in bold).}
\label{table_gart_shift}
\resizebox{\textwidth}{!}{
\begin{tabular}{lcccccccccccc}
\toprule
& \multicolumn{4}{c}{LPIPS $\downarrow$}
& \multicolumn{4}{c}{PSNR $\uparrow$}
& \multicolumn{4}{c}{SSIM $\uparrow$} \\
\cmidrule(lr){2-5}
\cmidrule(lr){6-9}
\cmidrule(lr){10-13}
Dog & Pixel & Ours & PoG & DT & Pixel & Ours & PoG & DT & Pixel & Ours & PoG & DT \\
\midrule
Alaskan & 0.2875 & \textbf{0.2664} & 0.3120 & 0.2847 &
20.0056 & \textbf{20.6333} & 16.5123 & 19.1629 &
0.8793 & \textbf{0.8845} & 0.8501 & 0.8745 \\

Shiba & 0.2749 & \textbf{0.1788} & 0.1805 & 0.1807 &
20.8241 & \textbf{25.3568} & 23.9165 & 24.7880 &
0.9069 & \textbf{0.9344} & 0.9319 & 0.9301 \\

Hound & 0.3406 & \textbf{0.2514} & 0.3361 & 0.2781 &
16.2762 & \textbf{19.4494} & 13.5665 & 17.8886 &
0.8372 & \textbf{0.8769} & 0.8202 & 0.8645 \\

Corgi & 0.1164 & 0.1100 & 0.1137 & \textbf{0.1082} &
25.4472 & \textbf{26.5250} & 24.8425 & 25.8148 &
0.9497 & \textbf{0.9561} & 0.9508 & 0.9533 \\

French & 0.3038 & \textbf{0.2339} & 0.2820 & 0.2436 &
17.6107 & \textbf{20.8778} & 17.0710 & 20.2074 &
0.8888 & \textbf{0.9106} & 0.8835 & 0.9064 \\

English & \textbf{0.2367} & 0.2418 & 0.4253 & 0.2408 &
21.2707 & \textbf{21.3255} & 8.9685 & 20.4019 &
\textbf{0.8939} & 0.8938 & 0.7494 & 0.8871 \\

Pitbull & 0.2505 & \textbf{0.2340} & 0.2881 & 0.2440 &
19.6348 & \textbf{20.2401} & 14.7755 & 18.1753 &
0.8851 & \textbf{0.8937} & 0.8514 & 0.8811 \\

\midrule
\textbf{Mean} &
0.2586 & \textbf{0.2166} & 0.2768 & 0.2257 &
20.1528 & \textbf{22.0583} & 17.0933 & 20.9199 &
0.8915 & \textbf{0.9071} & 0.8625 & 0.8996 \\
\bottomrule
\end{tabular}
}
\end{table*}

\subsection{Fast Motion Optimization}
Beyond the initial misalignment setting, we demonstrate our method's performance on a fast-motion scenario where the first frame is already aligned with the rendered object. We use a real-world video from~\cite{luiten2024dynamic}, optimizing with monocular supervision. We further increase the difficulty by temporally subsampling input frames. To apply our method, we separate foreground and background Gaussians, optimize only the foreground Gaussians, and recombine them at rendering time.

\begin{figure}[t]
  \centering
  \includegraphics[width=0.85\linewidth]{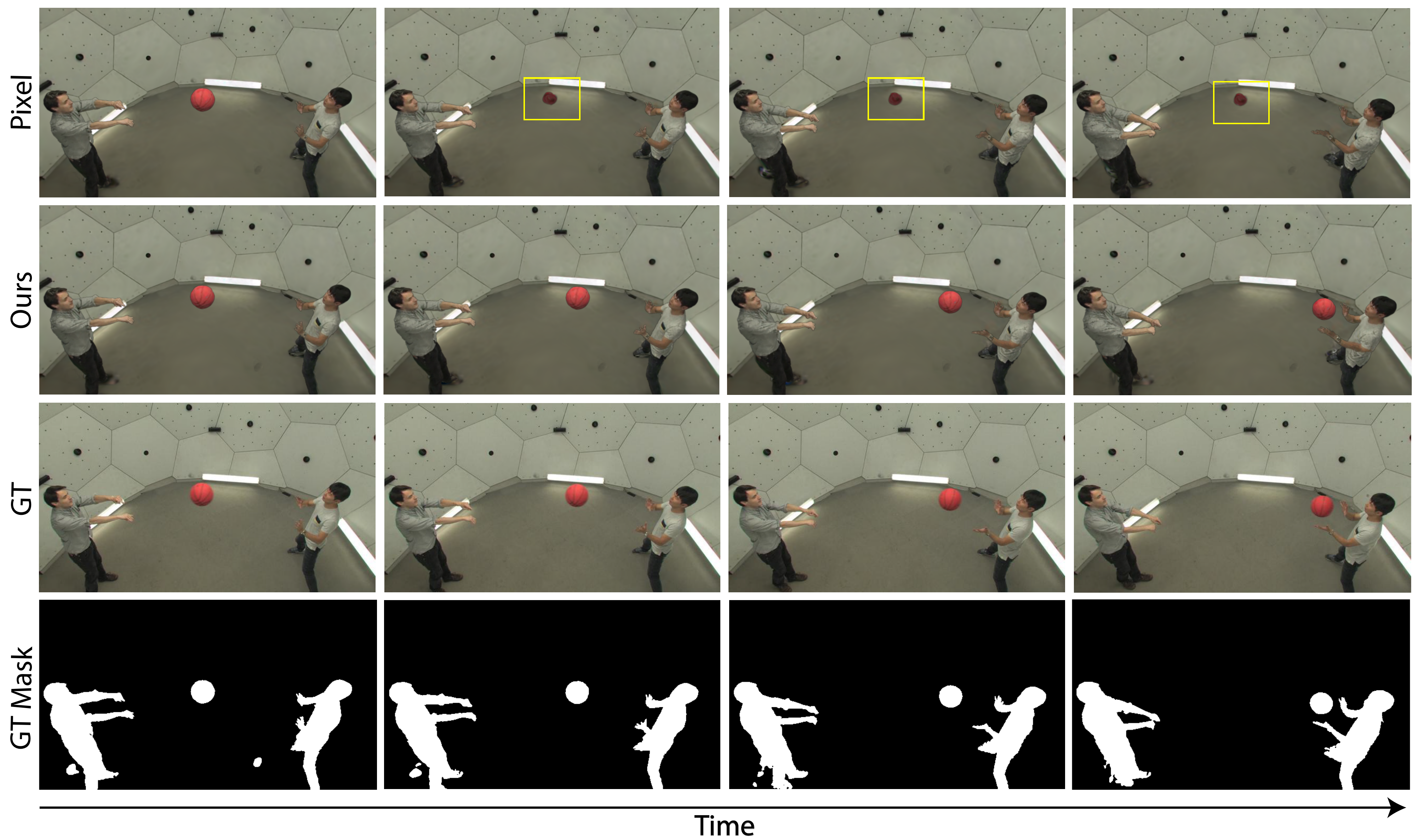}
  \caption{Qualitative comparison on a thrown basketball sequence from~\cite{luiten2024dynamic}. Large inter-frame motion creates a zero-overlap regime where pixel-based supervision fails, while our method recovers the correct alignment despite imperfect dataset masks (bottom).}
  \label{fig:new_data}
\end{figure}

Figure~\ref{fig:new_data} shows the ground-truth frames and corresponding masks, together with the results obtained using our method and pixel-based supervision. We note that the dataset-provided masks contain artifacts. Despite the correct first-frame alignment, the thrown basketball undergoes large motion between frames, resulting in a zero-overlap regime. In this scenario, pixel-based supervision fails, whereas our method maintains accurate alignment despite the imperfect masks.

\section{Conclusion}

We presented \shortname, a robust, model-agnostic framework that resolves the vanishing gradient problem inherent to dynamic 3DGS tracking. By replacing localized spatial losses with Spectral Moment supervision and a principled frequency annealing schedule, we allow tracking pipelines to recover from extreme spatial misalignments at initialization, bypassing the need for manual alignment or restrictive category-specific priors.

While \shortname significantly expands the basin of attraction, its current formulation assumes access to a pre-initialized canonical asset, restricting its scope to model-based tracking. A compelling future direction is extending this frequency-guided optimization beyond tracking to full dynamic scene reconstruction, where canonical geometry and motion are jointly optimized from uncalibrated video. Additionally, exploring alternative moment types to capture highly complex dynamics remains an exciting avenue for future research.

\section*{Acknowledgments}
We thank Gal Harari and Ido Sobol for their help with code and visualization, and Matan Atzmon for insightful discussions.
Avigail Cohen Rimon acknowledges support of the Miriam and Aaron Gutwirth Memorial Fellowship.
Mirela Ben Chen acknowledges support from the Israel Science Foundation (grant No. 1073/21).
Or Litany acknowledges support from the Israel Science Foundation (grant No. 624/25) and the Azrieli Foundation Early Career Faculty Fellowship. This research was supported by the Council for Higher Education in Israel under the Moonshot Project.

\bibliographystyle{splncs04}
\bibliography{main}

\clearpage
\appendix

\addtocontents{toc}{\protect\setcounter{tocdepth}{2}}

\title{SpectralSplats: Robust Differentiable Tracking... - Supplementary Material}
\titlerunning{SpectralSplats}

\author{Avigail Cohen Rimon \inst{1} \and ... }
\authorrunning{SpectralSplats}

\institute{Technion - Israel Institute of Technology\and Nvidia}



\newpage

\section{Derivation of the Phase-Wrapping Condition}
\label{sec:derivation}

In this section we provide the detailed derivation of the condition $|\bm{\omega}^T\mathbf{d}| < \pi$ stated in Sec.~3.3 of the main paper, which ensures that the spectral loss at frequency $\bm{\omega}$ possesses a unique basin of attraction with monotonically directed gradients toward the correct solution. 

\paragraph{Setup.}
Consider a rendered image $\mathbf{I}_{\mathrm{rend}}$ that is a spatially displaced copy of the ground-truth target $\mathbf{I}_{\mathrm{gt}}$, i.e.\ $\mathbf{I}_{\mathrm{rend}}(\mathbf{p}) = \mathbf{I}_{\mathrm{gt}}(\mathbf{p} - \mathbf{d})$, where $\mathbf{d} \in \mathbb{R}^2$ denotes the spatial misalignment vector we wish to drive to zero.
For a discrete 2D frequency vector $\bm{\omega}$, we defined the spectral moment of an image $\mathbf{I}$ as
\begin{equation}
  \mathcal{M}(\bm{\omega};\,\mathbf{I})
  \;=\;
  \sum_{\mathbf{p}} \mathbf{I}(\mathbf{p})\,\exp\!\bigl(-j\,\bm{\omega}^T\mathbf{p}\bigr).
  \label{eq:spectral_moment}
\end{equation}
By the Fourier shift theorem, a spatial translation by $\mathbf{d}$ maps to a phase shift in the frequency domain:
\begin{equation}
  \mathcal{M}(\bm{\omega};\,\mathbf{I}_{\mathrm{rend}})
  \;=\;
  \mathcal{M}(\bm{\omega};\,\mathbf{I}_{\mathrm{gt}})\;\exp\!\bigl(-j\,\bm{\omega}^T\mathbf{d}\bigr).
  \label{eq:shift_theorem}
\end{equation}

\paragraph{Single-frequency spectral loss.}
We define the spectral loss at frequency $\bm{\omega}$ as the squared magnitude of the difference between the rendered and target moments:
\begin{equation}
  E(\mathbf{d};\,\bm{\omega})
  \;=\;
  \tfrac{1}{2}\,
  \bigl|\mathcal{M}(\bm{\omega};\,\mathbf{I}_{\mathrm{rend}})
        - \mathcal{M}(\bm{\omega};\,\mathbf{I}_{\mathrm{gt}})\bigr|^2.
  \label{eq:spectral_loss}
\end{equation}
Substituting Eq.~\eqref{eq:shift_theorem} and letting $\mathcal{M}_{\mathrm{gt}} \equiv \mathcal{M}(\bm{\omega};\,\mathbf{I}_{\mathrm{gt}})$, we obtain
\begin{equation}
  E(\mathbf{d};\,\bm{\omega})
  \;=\;
  \tfrac{1}{2}\,|\mathcal{M}_{\mathrm{gt}}|^2\,
  \bigl|e^{-j\,\bm{\omega}^T\mathbf{d}} - 1\bigr|^2.
  \label{eq:spectral_loss_shifted}
\end{equation}
We now expand the squared complex magnitude.  Denoting $\phi = \bm{\omega}^T\mathbf{d}$,
\begin{equation}
  \bigl|e^{-j\phi} - 1\bigr|^2
  \;=\;
  2 - 2\cos\phi,
  \label{eq:trig_identity}
\end{equation}
so the spectral loss reduces to the compact form
\begin{equation}
  \boxed{
  E(\mathbf{d};\,\bm{\omega})
  \;=\;
  |\mathcal{M}_{\mathrm{gt}}|^2\,
  \bigl(1 - \cos(\bm{\omega}^T\mathbf{d})\bigr).}
  \label{eq:cosine_loss}
\end{equation}

\paragraph{Phase-wrapping condition and the basin of attraction.}
Differentiating Eq.~\eqref{eq:cosine_loss} with respect to $\mathbf{d}$ yields the gradient
\begin{equation}
  \nabla_{\mathbf{d}}\,E
  \;=\;
  |\mathcal{M}_{\mathrm{gt}}|^2\,
  \sin(\bm{\omega}^T\mathbf{d})\;\bm{\omega}.
  \label{eq:gradient}
\end{equation}
Unlike the standard spatial cross-term analysed in Sec.~3.1, this gradient is non-zero whenever $\bm{\omega}^T\mathbf{d} \neq n\pi$, $n\in\mathbb{Z}$, confirming that the spectral objective provides a valid, directional signal even when the rendered and target images are spatially disjoint.  The key question is: \emph{from what range of initial displacements does the unique global minimum at $\mathbf{d}=\mathbf{0}$ remain the sole attractor?}  The stationary points of $E$ satisfy $\sin(\bm{\omega}^T\mathbf{d}) = 0$, i.e.\ $\bm{\omega}^T\mathbf{d} = n\pi$.  Among these, $n=0$ is the global minimum ($E=0$), the odd multiples $n = \pm 1, \pm 3, \ldots$ are local maxima of the $1-\cos$ profile, and the even multiples $n = \pm 2, \pm 4, \ldots$ are \emph{false} global minima where $E=0$ but $\mathbf{d}\neq\mathbf{0}$.  Crucially, the function $1-\cos(\phi)$ is \emph{strictly monotonically increasing} on $(0, \pi)$ and strictly decreasing on $(-\pi, 0)$.  Therefore, any gradient-based optimiser initialised with a displacement satisfying $|\bm{\omega}^T\mathbf{d}| < \pi$ will follow a monotonically descending path toward $\mathbf{d}=\mathbf{0}$ without encountering any intervening stationary point.  Once the phase exceeds $\pi$, the loss begins to decrease toward the \emph{next} period's minimum at $\bm{\omega}^T\mathbf{d} = 2\pi$, creating a false basin that traps the optimiser at an incorrect alignment.  This establishes the phase-wrapping condition:
\begin{equation}
  \boxed{|\bm{\omega}^T\mathbf{d}_t| \;<\; \pi}
  \label{eq:phase_wrap}
\end{equation}
as the necessary and sufficient condition for the spectral loss at frequency $\bm{\omega}$ to provide a unique, correct basin of attraction at optimisation step $t$.

\paragraph{Quadratic regime and exponential convergence.}
When the phase-wrapping condition is satisfied, the Taylor expansion $1 - \cos(\phi) \approx \phi^2/2$ yields the quadratic approximation
\begin{equation}
  E(\mathbf{d};\,\bm{\omega})
  \;\approx\;
  \tfrac{1}{2}\,|\mathcal{M}_{\mathrm{gt}}|^2\,(\bm{\omega}^T\mathbf{d})^2,
  \label{eq:quadratic}
\end{equation}
with gradient $\nabla_{\mathbf{d}} E \approx |\mathcal{M}_{\mathrm{gt}}|^2\,(\bm{\omega}^T\mathbf{d})\,\bm{\omega} \propto \mathbf{d}$.  
Gradient descent would thus follow exponential convergence, since:
\begin{equation}
  \mathbf{d}_{t+1}
  \;=\;
  \mathbf{d_t} - \eta\,|\mathcal{M}_{\mathrm{gt}}|^2\,\|\bm{\omega}\|^2\, \mathbf{d}_t
  \;=\;
  \underbrace{\bigl(1 - \eta\,|\mathcal{M}_{\mathrm{gt}}|^2\,\|\bm{\omega}\|^2\bigr)}_{\displaystyle\gamma}\; \mathbf{d}_t.
  \label{eq:gd_step}
\end{equation}
For a sufficiently small learning rate $\eta$ the contraction factor $\gamma$ lies in $(0,1)$.  Unrolling the recursion yields exponential decay of the spatial displacement:
\begin{equation}
  \mathbf{d}_t \;=\; \mathbf{d}_0\,\gamma^t
  \qquad\text{with } \gamma \in (0,1).
  \label{eq:exp_convergence}
\end{equation}

\paragraph{From exponential convergence to the linear annealing schedule.}
Combining Eqs.~\eqref{eq:phase_wrap} and~\eqref{eq:exp_convergence}, the maximum safe frequency magnitude at step $t$ must satisfy
\begin{equation}
  \|\bm{\omega}_{\max}(t)\|
  \;<\;
  \frac{\pi}{\|\mathbf{d}_t\|}
  \;\leq\;
  \frac{\pi}{\|\mathbf{d}_0\|\,\gamma^t}
  \;\propto\;
  \gamma^{-t}.
  \label{eq:freq_bound}
\end{equation}
On a standard spectral grid the discrete frequencies are organised logarithmically, $\|\bm{\omega}_k\| \propto 2^k$, so matching the exponential growth $\gamma^{-t}$ to $2^{k(t)}$ gives
\begin{equation}
  2^{k(t)} \;\propto\; \gamma^{-t}
  \quad\Longrightarrow\quad
  k(t) \;=\; \frac{t\,\log(1/\gamma)}{\log 2}
  \;\propto\; t.
  \label{eq:linear_schedule}
\end{equation}
That is, the active frequency \emph{index} must grow linearly with the optimisation step.  This provides a rigorous, first-principles justification for the linear annealing schedule $\alpha(t)$ that scales from $0$ to $K$ over the course of optimisation, as stated in Eq.~(7) of the main paper.  In summary: exponential decay of the spatial error permits exponential growth of the safe frequency bandwidth, which, on a logarithmic frequency grid, translates to a strictly linear expansion of the active frequency index --- ensuring the optimiser remains within the unique basin of attraction at every step while progressively recovering fine spatial detail.

\clearpage
\section{Demo}
\begin{figure}[h]
    \centering
    \includegraphics[width=\textwidth]{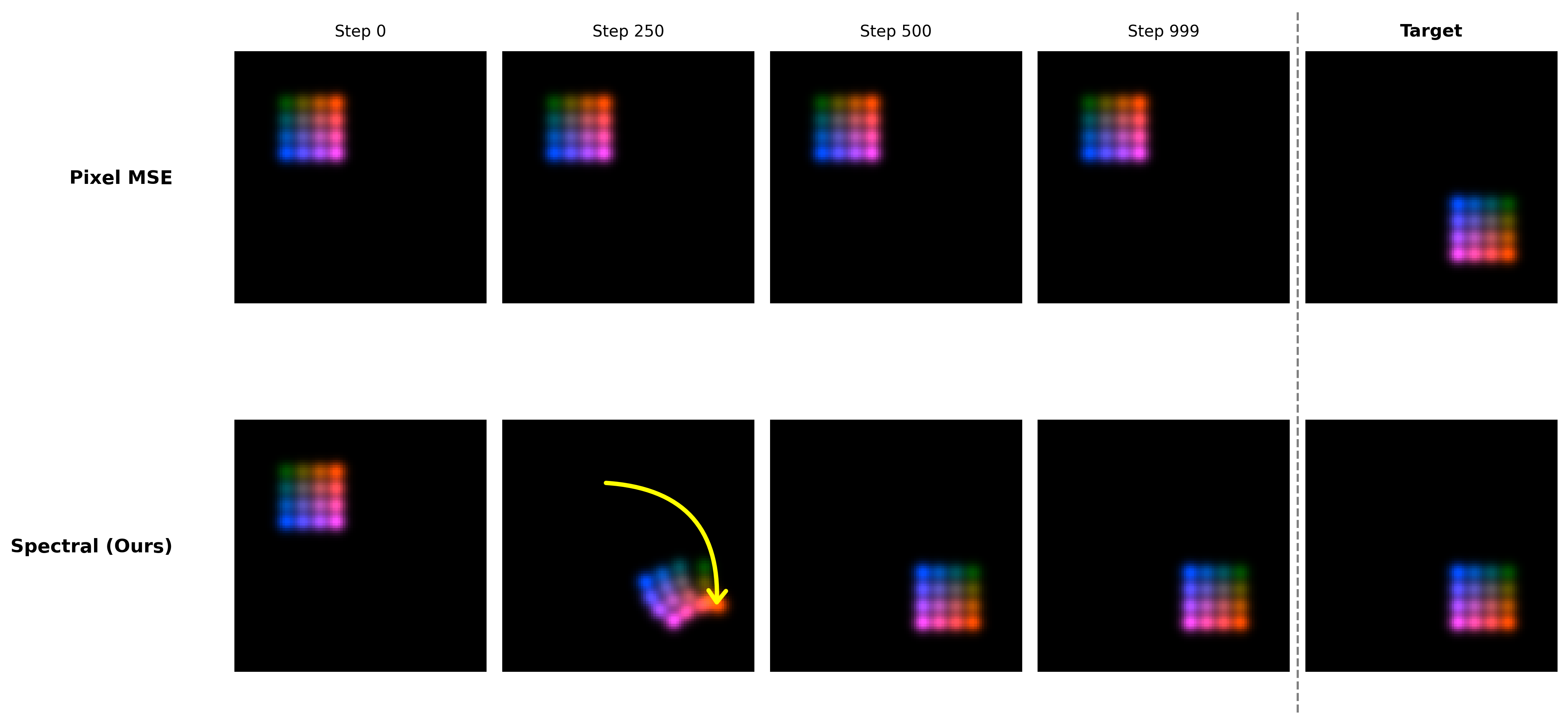}
    \caption{2D optimization demo under large spatial misalignment (translation and rotation). Pixel MSE supervision (top) fails to move toward the target, remaining near initialization. Spectral supervision (bottom) produces coherent global motion and successfully converges to the target.}
    \label{fig:2d_demo_comparison}
\end{figure}

\begin{wrapfigure}{r}{0.5\textwidth}
    \centering
    \includegraphics[width=0.9\linewidth]{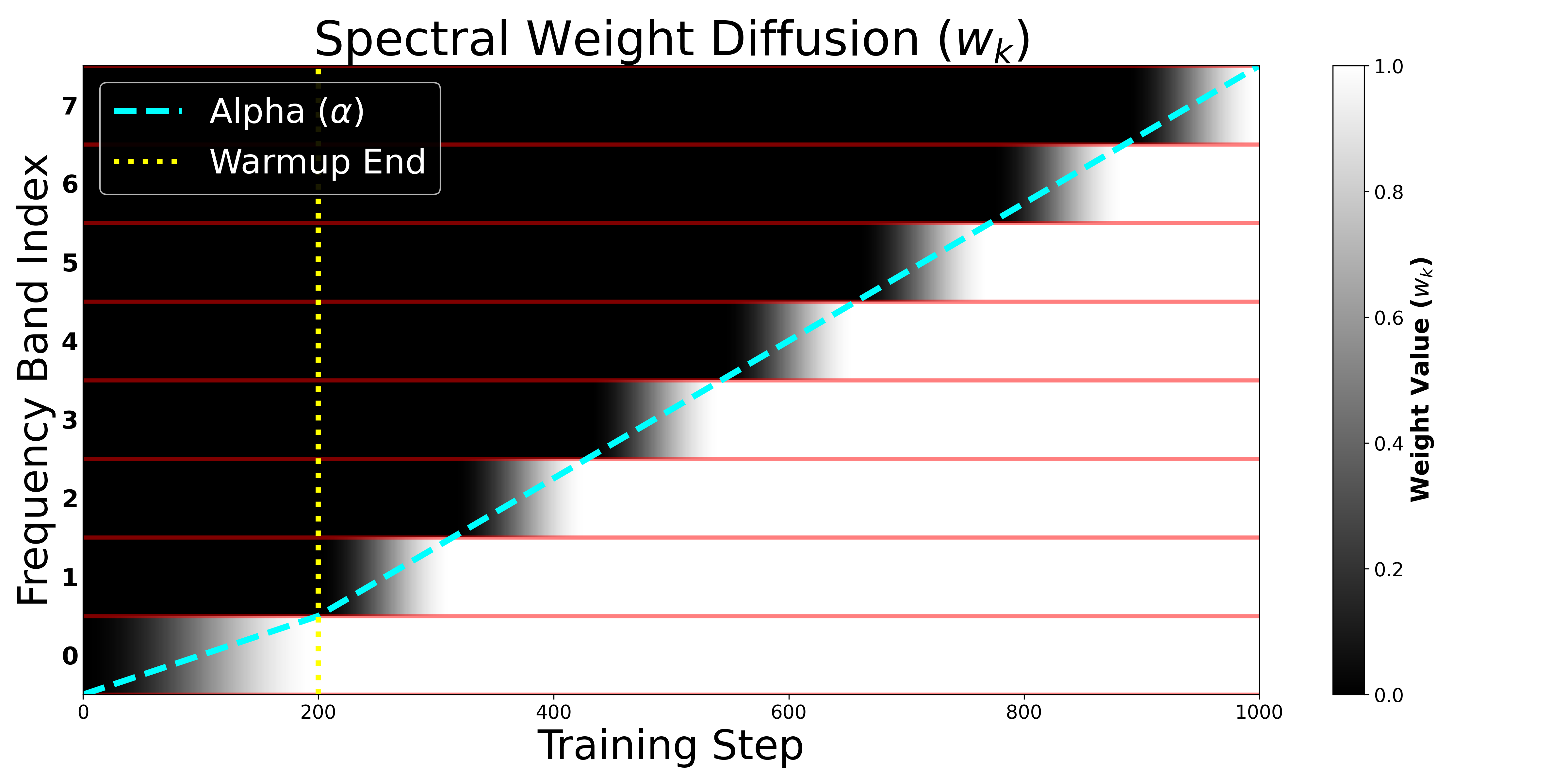}
    \caption{Visualization of the frequency annealing schedule. After an initial warm-up phase, the active bandwidth $\alpha(t)$ increases linearly, and higher frequencies gradually fade in via the cosine weighting $w_k(t)$, enabling a smooth transition from global alignment to fine-grained refinement.} 
    \label{fig:2d_demo_weights}
\end{wrapfigure}
To further build intuition, we provide the code for illustrative 1D and 2D demos that visualize the optimization with our method. The code is included with the supplementary material.
The 1D demo is shown in Fig.~2 of the Method section. Figure~\ref{fig:2d_demo_comparison} presents the 2D demo, where the source and target exhibit large spatial misalignment involving both translation and rotation. The top row (Pixel MSE) shows that pixel supervision fails to recover alignment: the pattern remains stranded near initialization. In contrast, the bottom row (Spectral) demonstrates a coherent motion toward the target (rightmost column), illustrating how Spectral Moment supervision establishes a global basin of attraction and successfully resolves the displacement. 

Figure~\ref{fig:2d_demo_weights} visualizes the annealing schedule. During the initial warm-up phase (left of the dashed line), only the lowest frequency band is active. $\alpha(t)$ then increases linearly, and higher frequencies gradually fade in via the cosine weighting $w_k(t)$. Early low-frequency dominance promotes global convergence by first resolving coarse misalignments, such as global translation and rotation (as observed around step 250 in Fig~\ref{fig:2d_demo_comparison}). As additional frequencies become active, finer geometric details are refined (e.g., around step 500).

\clearpage

\section{Training Objective}
Tracking with differentiable 3D Gaussian Splatting is an active research area, with recent works proposing diverse deformation parameterizations and regularization strategies to improve stability and convergence.
In this work, we build upon GSGD~\cite{bekor2025gaussian} as a representative state-of-the-art tracking framework, adopting its deformation parameterization and regularization terms.

Consistent with the Method section of the paper, we adopt a two-phase optimization scheme. We first optimize a spectrally annealed objective to establish a global basin of attraction. 
Following that, we transition to spatial-domain supervision for high-frequency refinement.
In alignment with the notation introduced in Section~3 of the paper, let $I_{\text{rend}}(\mathbf{p};\Theta) = \mathcal{R}(\mathcal{D}(\mathcal{G}_{\text{ref}};\Theta))$ denote the rendered RGB image and $O_{\text{rend}}$ its rendered opacity map. Let $I_{\text{gt}}$ and $O_{\text{gt}}$ denote the target RGB image and mask.

\subsection{Loss Components}
\paragraph{Spectral Phase.}
We define the spectral moment for a 2D spatial frequency vector $\omega_{k_x, k_y}$ using a phase-scaling factor $0.5\pi$:
\begin{equation}
\mathcal{M}(k_x, k_y; I) = \sum_{\mathbf{p}} I(\mathbf{p}) \exp(j \cdot 0.5\pi \cdot \omega_{k_x, k_y}^\top \mathbf{p})
\end{equation}
During the spectral stage, we supervise the current pose by minimizing the discrepancy between rendered and target spectral signatures over the active frequency band $\mathcal{K}(t)$:
\begin{equation}
\begin{split}
\mathcal{L}_{\text{image}}^{\text{spectral}} = &\sum_{k \in \mathcal{K}(t)} w_k(t) \left\| \mathcal{M}_k(I_{\text{rend}}) - \mathcal{M}_k(I_{\text{gt}}) \right\|_1 \\
&+ \lambda_{\text{mask}} \sum_{k \in \mathcal{K}(t)} w_k(t) \left\| \mathcal{M}_k(O_{\text{rend}}) - \mathcal{M}_k(O_{\text{gt}}) \right\|_1
\end{split}
\end{equation}

\paragraph{Spatial (Pixel) Phase.}
The spatial objective is defined as:
\begin{equation}
\mathcal{L}_{\text{image}}^{\text{pixel}} = \| I_{\text{rend}} - I_{\text{gt}} \|_2^2 + \| I_{\text{rend}} \odot O_{\text{rend}} - I_{\text{gt}} \odot O_{\text{gt}} \|_2^2 + \lambda_{bce} \mathrm{BCE}(O_{\text{rend}}, O_{\text{gt}})
\end{equation}
where $\odot$ denotes element-wise multiplication. In our SC4D experiments, we alternatively replace $\mathcal{L}_{\text{image}}^{\text{pixel}}$ with LPIPS$(I_{rend}, I_{gt})$ supervision during this phase.

\paragraph{Overall Objective.}
The total loss minimized during optimization is:
\begin{equation}
\mathcal{L} = \lambda_{\text{image}} \mathcal{L}_{\text{image}} + \lambda_{\text{arap}} \mathcal{E}_{\text{arap}}
\end{equation}
where $\mathcal{L}_{\text{image}}$ corresponds to either the spectral or spatial formulation depending on the training phase. Following GSGD, we apply As-Rigid-As-Possible (ARAP) regularization to encourage locally rigid motion of control points.

\subsection{Spatial Loss Ablation}
We ablate the different loss components of the spatial-phase loss $\mathcal{L}_{image}^{pixel}$ and report PSNR results for a representative example from GART (Shiba) under the 0.6 shift setting. We compare MLP+Ours and MLP+Pixel performance under the different loss variants. Overall, the full version achieves the best performance. While the improvement is not substantial, it consistently provides the strongest results, suggesting that each component contributes to fine-tuning the final outcome.
As noted, for SC4D we also report results where $\mathcal{L}_{image}^{pixel}$ is replaced with an LPIPS loss, further demonstrating the effectiveness of our method regardless of the choice of spatial loss.
\begin{table}[h]
\caption{Ablation study of the loss components in $\mathcal{L}_{\text{image}}^{\text{pixel}}$. PSNR is reported for each configuration.}
\centering
\small
\setlength{\tabcolsep}{8pt}
\renewcommand{\arraystretch}{1.1}
\begin{tabular}{l c c c c}
\hline
\textbf{Method} & \textbf{MSE} & \textbf{MSE + Masked MSE} & \textbf{MSE + BCE} & \textbf{All} \\
\hline
MLP+Ours & 25.011 & 25.122 & 25.295 & \textbf{25.356} \\
MLP+Pixel    & 19.454 & 19.747 & 19.237 & \textbf{20.824} \\
\hline
\end{tabular}
\label{tab:loss_ablation}
\end{table}

\begin{table}[h]
\caption{Ablation study of loss components in $\mathcal{L}_{\text{image}}^{\text{pixel}}$ across the GART dataset. PSNR is reported for each configuration.}
\centering
\small
\setlength{\tabcolsep}{8pt}
\renewcommand{\arraystretch}{1.1}
\begin{tabular}{l c c c c}
\hline
\textbf{Method} & \textbf{MSE} & \textbf{MSE + Masked MSE} & \textbf{MSE + BCE} & \textbf{All} \\
\hline
MLP+Ours & 20.651 & 20.206 & 20.553 & \textbf{22.058} \\
MLP+Pixel    & 16.448 & 16.139 & 20.003 & \textbf{20.152} \\
\hline
\end{tabular}
\label{tab:psnr_ablation}
\end{table}

\subsection{Annealing Schedule Ablation}
We ablate the annealing schedule $\alpha(t)=tK$, where $t$ denotes the normalized optimization step and $K$ the number of frequency bands. On GART with a shift radius of $0.6$, we evaluate $K\in\{4,6,8,10,12\}$ and obtain PSNR values of $(21.79, 22.02, 22.06, 21.28, 21.35)$, respectively. We observe that the method is relatively stable across a broad range of schedules, with the best result achieved at $K=8$.

\clearpage
\section{SC4D Experiment}
\subsection{Performance under Aligned Initialization}
To further evaluate the robustness of our method, we analyze the SC4D experiment in the aligned setting (shift = 0.0), where the initial pose is set to the first-frame pose provided by SC4D and therefore matches the supervision. This experiment verifies that our method does not degrade performance when no initial misalignment is present.
Similar to Table~1 in the paper, which reports results under shift = 0.5, Table~\ref{tab:sc4d_shift0} presents quantitative results across different deformation parameterizations and supervision variants.

We observe that our method consistently matches or outperforms pixel-only supervision across PSNR, SSIM, and LPIPS, on both training and novel views. These results confirm that Spectral Moment supervision is not only robust under severe misalignment, but also remains beneficial - or at worst neutral, when the initialization is well aligned.
\begin{table}[h]
\centering
\caption{Evaluation of our method on the synthetic SC4D dataset with shift = 0.0, i.e., when the 3DGS model is initially aligned with the supervision. Our method does not degrade performance and improves results in most cases. This highlights its robustness, as it enhances stability without compromising performance in settings where pixel-only supervision does not exhibit catastrophic failure.}
\label{tab:sc4d_shift0}
\resizebox{\textwidth}{!}{%
\begin{tabular}{l|l|ccc|ccc}
\toprule
& Loss & LPIPS $\downarrow$ & PSNR $\uparrow$ & SSIM $\uparrow$ & NV-LPIPS $\downarrow$ & NV-PSNR $\uparrow$ & NV-SSIM $\uparrow$ \\
\midrule
\multirow{2}{*}{MLP w. LPIPS}
& Pixel & 0.0403 & 27.3404 & 0.9591 & 0.0914 & 19.3308 & \textbf{0.9344} \\
& Ours & \textbf{0.0325} & \textbf{29.3938} & \textbf{0.9636} & \textbf{0.0870} & \textbf{19.3311} & 0.9312 \\
\midrule
\multirow{2}{*}{MLP w/o LPIPS}
& Pixel & 0.0905 & 23.7205 & 0.9418 & 0.1490 & 16.2879 & 0.9182 \\
& Ours & \textbf{0.0472} & \textbf{28.3403} & \textbf{0.9594} & \textbf{0.1255} & \textbf{17.9528} & \textbf{0.9214} \\
\midrule
\multirow{2}{*}{Direct w. LPIPS}
& Pixel & 0.1271 & 18.1281 & 0.9127 & \textbf{0.0965} & \textbf{18.8798} & \textbf{0.9343} \\
& Ours & \textbf{0.1046} & \textbf{20.1747} & \textbf{0.9255} & 0.0998 & 18.7325 & 0.9319 \\
\midrule
\multirow{2}{*}{Direct w/o LPIPS}
& Pixel & 0.0959 & 22.0955 & 0.9389 & 0.0978 & 19.0892 & 0.9331 \\
& Ours & \textbf{0.0940} & \textbf{22.2446} & \textbf{0.9406} & \textbf{0.0944} & \textbf{19.3347} & \textbf{0.9355} \\
\bottomrule
\end{tabular}
}
\end{table}

\subsection{More Qualitative Results}
\begin{figure}[t]
  \centering
  \includegraphics[width=\textwidth]{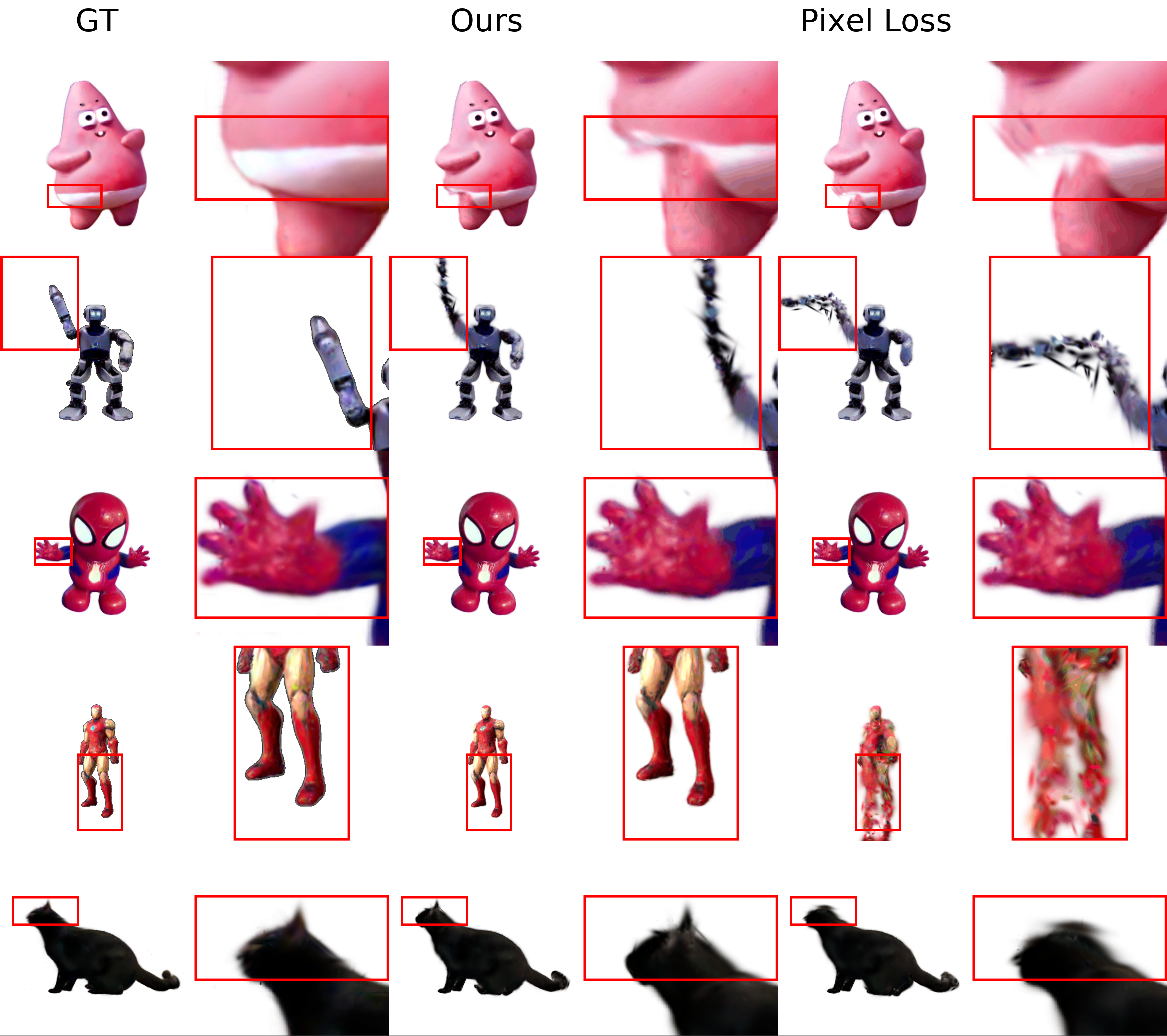}
  \caption{Qualitative comparison of our method against pixel loss optimization on the final frame without spatial misalignment. The pose initialization is set to the video's first frame.}
  \label{fig:comparison_shift0}
\end{figure}

\begin{figure}[t]
  \centering
  \includegraphics[width=\textwidth]{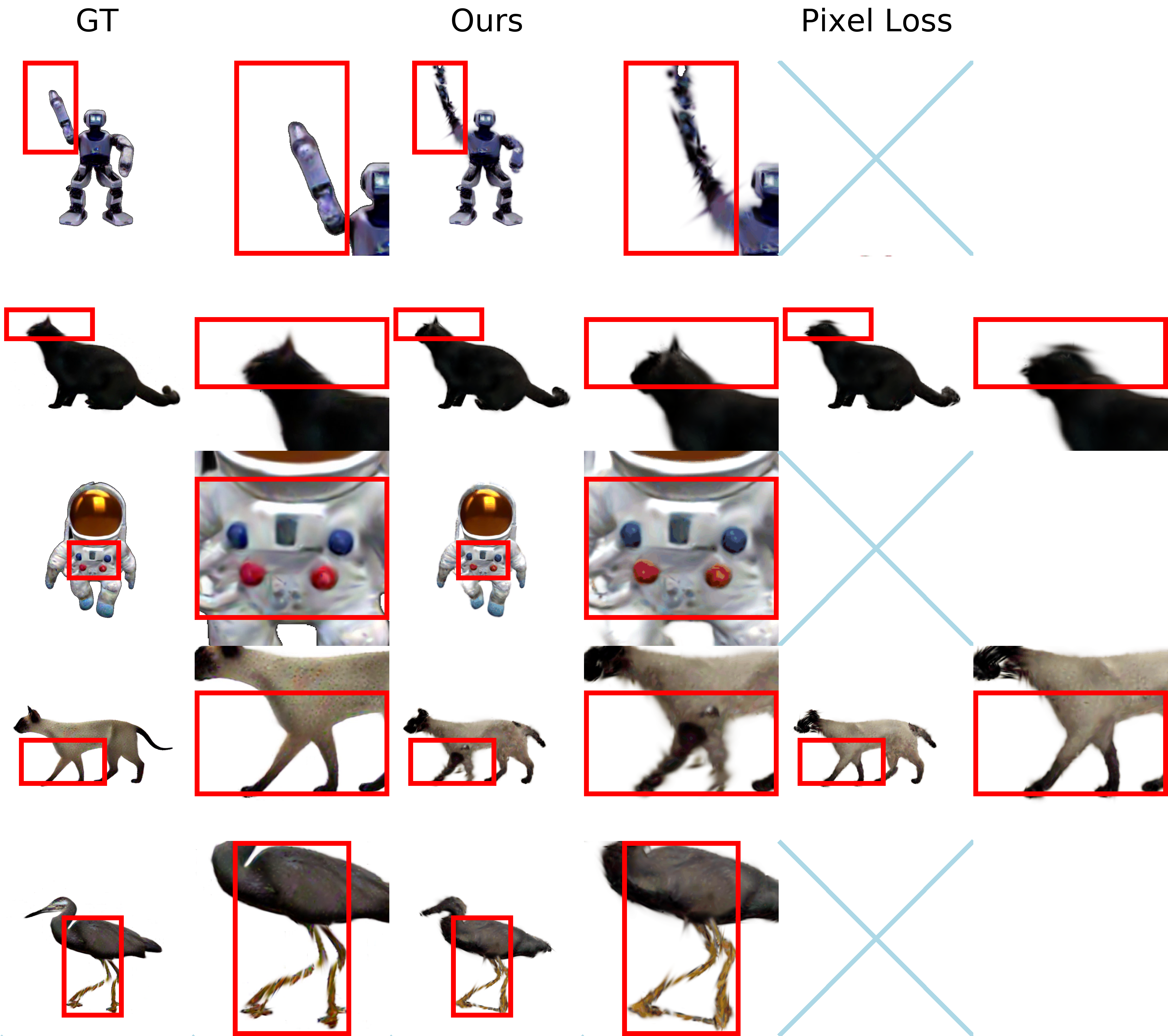}
  \caption{Qualitative comparison of our method against pixel loss optimization on the final frame with an initial pose offset of 0.5 from the video's first frame. Blue X marks empty frames where the optimization resulted with no gaussians present in the train view's frame.}
  \label{fig:comparison_shift0.5}
\end{figure}
Figures~\ref{fig:comparison_shift0} and~\ref{fig:comparison_shift0.5} present additional qualitative comparisons between our method and pixel-based optimization on SC4D.
In the aligned setting (shift = 0.0, Fig.~\ref{fig:comparison_shift0}), both methods recover the target pose; however, our method produces sharper details and cleaner structure, as seen in the regions highlighted by the red boxes.
Under spatial misalignment (shift = 0.5, Fig.~\ref{fig:comparison_shift0.5}), the difference becomes more pronounced. While our method maintains stable results, pixel-only optimization exhibits noticeable artifacts or collapses as the object drifts out of the frame (indicated by a blue × in the figure).

\subsection{Multi-View Training Analysis}
\begin{figure*}[t]
    \centering

    \begin{subfigure}{0.85\textwidth}
        \centering
        \includegraphics[width=\textwidth]{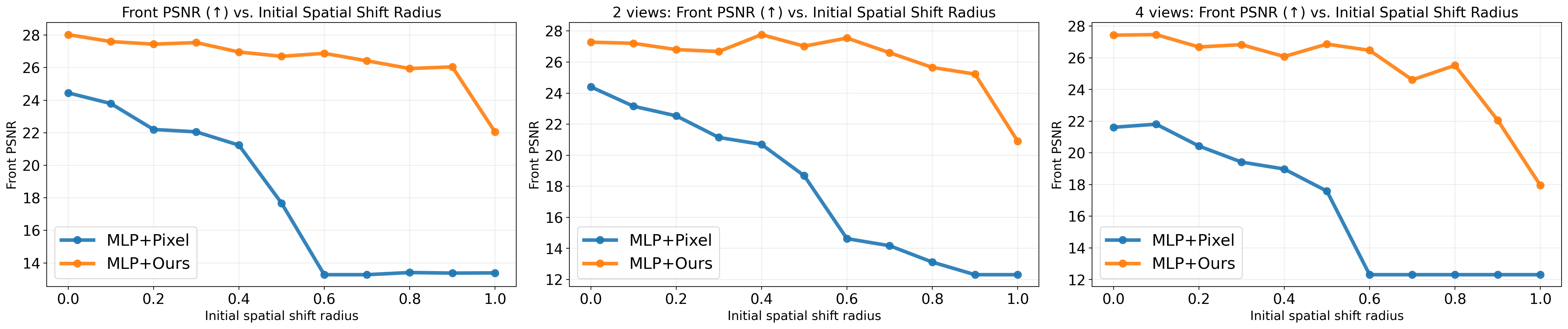}
        \caption{PSNR -- Front (training) view}
        \label{fig:sc4d_multiview_PSNR_front}
    \end{subfigure}

    \vspace{4pt}

    \begin{subfigure}{0.85\textwidth}
        \centering
        \includegraphics[width=\textwidth]{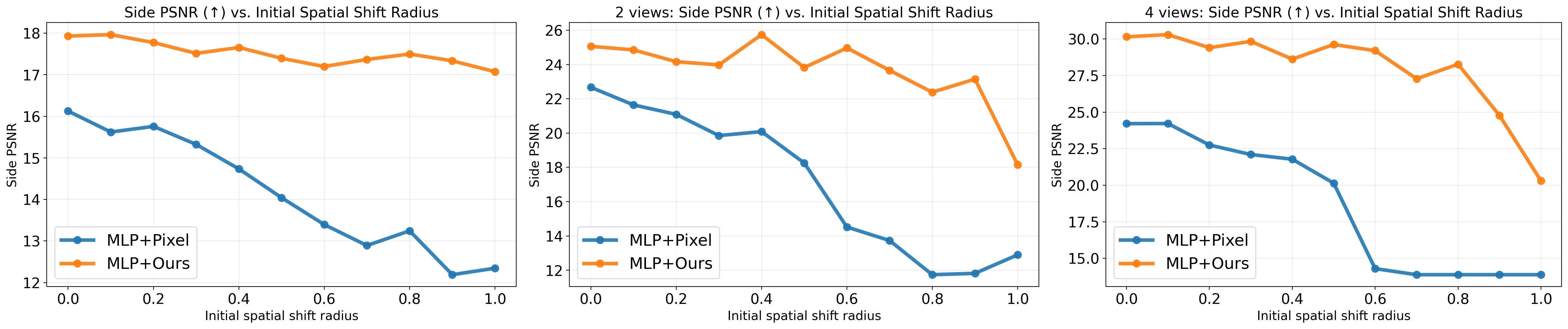}
        \caption{PSNR -- Side (novel) view}
        \label{fig:sc4d_multiview_PSNR_novel}
    \end{subfigure}

    \vspace{4pt}

    \begin{subfigure}{0.85\textwidth}
        \centering
        \includegraphics[width=\textwidth]{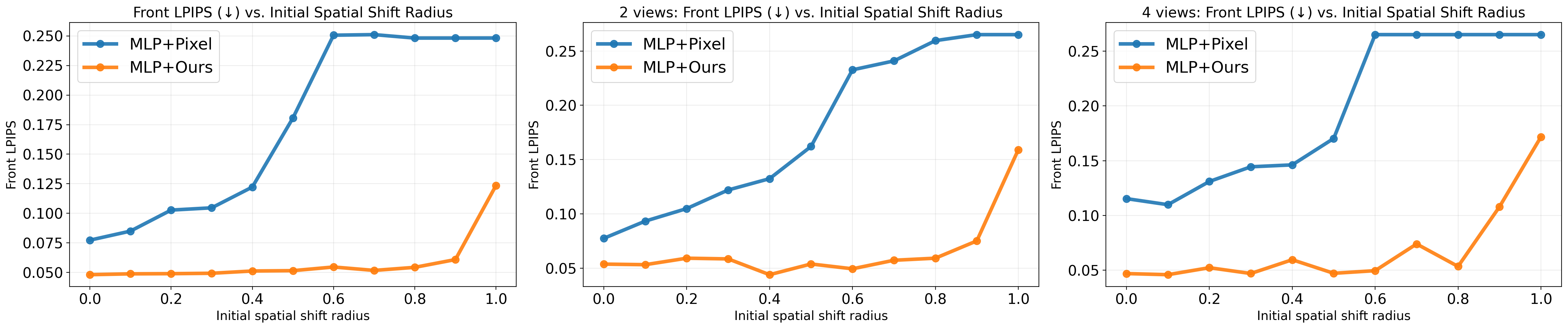}
        \caption{LPIPS -- Front (training) view}
        \label{fig:sc4d_multiview_LPIPS_front}
    \end{subfigure}

    \vspace{4pt}

    \begin{subfigure}{0.85\textwidth}
        \centering
        \includegraphics[width=\textwidth]{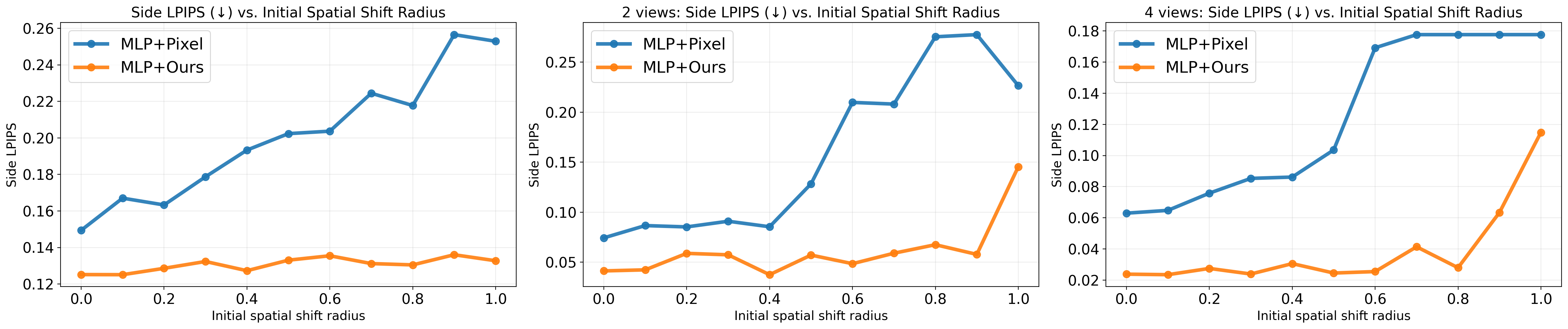}
        \caption{LPIPS -- Side (novel) view}
        \label{fig:sc4d_multiview_LPIPS_novel}
    \end{subfigure}

    \vspace{4pt}

    \begin{subfigure}{0.85\textwidth}
        \centering
        \includegraphics[width=\textwidth]{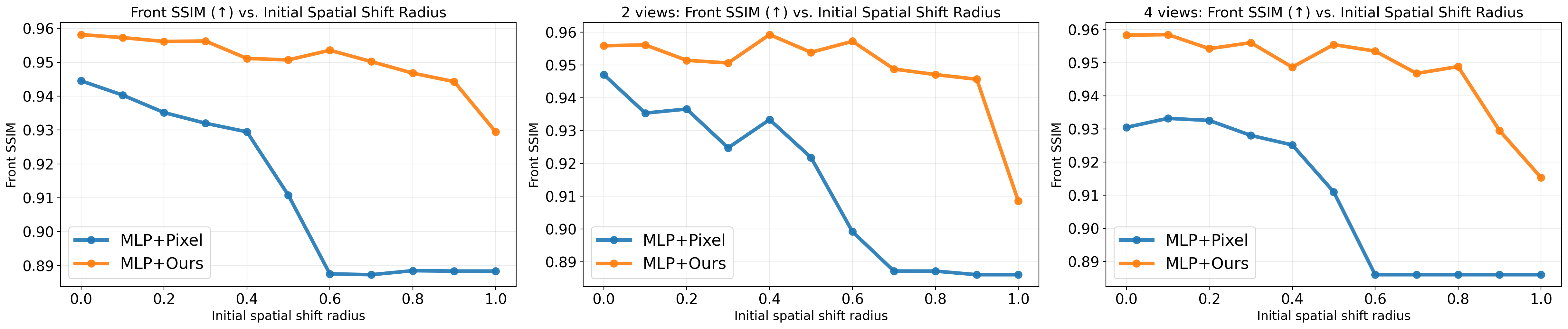}
        \caption{SSIM -- Front (training) view}
        \label{fig:sc4d_multiview_SSIM_front}
    \end{subfigure}

    \vspace{4pt}

    \begin{subfigure}{0.85\textwidth}
        \centering
        \includegraphics[width=\textwidth]{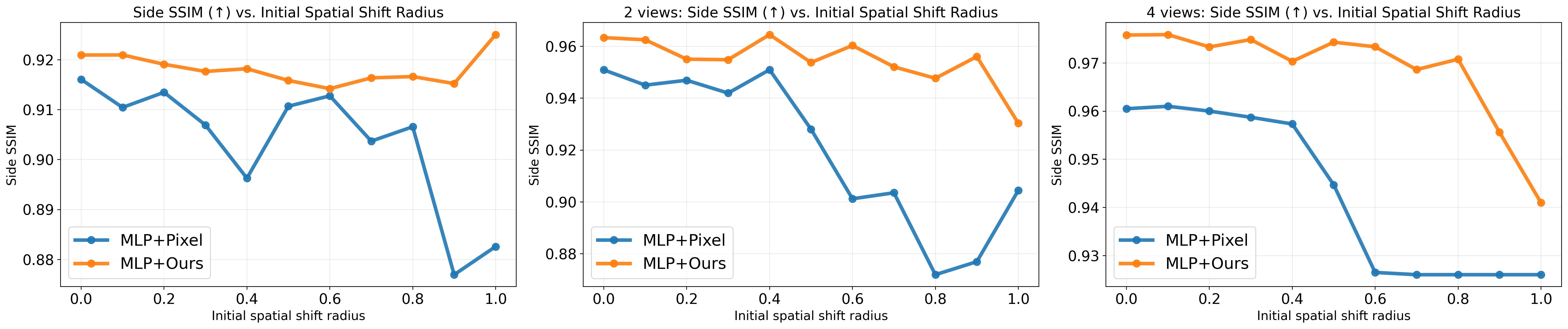}
        \caption{SSIM -- Side (novel) view}
        \label{fig:sc4d_multiview_SSIM_novel}
    \end{subfigure}

    \caption{Effect of multi-view supervision on SC4D under spatial misalignment. From top to bottom: PSNR (front), PSNR (side), LPIPS (front), LPIPS (side), SSIM (front), and SSIM (side). Across all view configurations, pixel-only supervision degrades under increasing shifts, whereas our method remains more stable, generalizes better to the novel view, and consistently achieves stronger performance.}
    \label{fig:sc4d_multiview_metrics_combined}
\end{figure*}
In the main paper, we report results using a single training view. Here, we further evaluate the effect of increasing the number of supervision views on the SC4D dataset.
Specifically, we compare training with one, two, and four views.
For the single-view setting, we use view angle $0^\circ$.
For two views, we supervise with $0^\circ$ and $180^\circ$.
For four views, we use $0^\circ$, $90^\circ$, $180^\circ$, and $270^\circ$.
We evaluate performance on both the \emph{front} view ($0^\circ$) and the \emph{side} view ($90^\circ$).

Figure~\ref{fig:sc4d_multiview_metrics_combined} reports PSNR, LPIPS and SSIM on the \emph{front} view and the \emph{side} view (novel for the single- and two-view settings), respectively, as a function of the initial shift radius.
Figure~\ref{fig:sc4d_multiview_PSNR_front} shows that across all view configurations, pixel-only supervision degrades rapidly as misalignment increases, whereas our method remains significantly more stable. Importantly, our approach achieves higher performance in all cases, including the zero-shift setting.
Figure~\ref{fig:sc4d_multiview_PSNR_novel} further demonstrates that although adding training views improves overall PSNR for both methods, pixel-based optimization remains sensitive to initialization and collapses under larger shifts. In contrast, our method consistently outperforms the baseline and exhibits more stable generalization across shifts and viewpoints.
LPIPS and SSIM exhibit trends consistent with PSNR, further confirming the robustness and superior generalization of our method.

\clearpage
\section{GART Experiment}
\subsection{Implementation Details}
\paragraph{Mask-Based Supervision.}
Given that the input videos contain background and scene context, we use the per-frame masks provided with the dataset to enable accurate supervision. The masks isolate the asset, and the rendered outputs are composited over a uniform background.

\paragraph{GART Initialization.}
\begin{figure}[h]
    \centering
    \includegraphics[width=\textwidth]{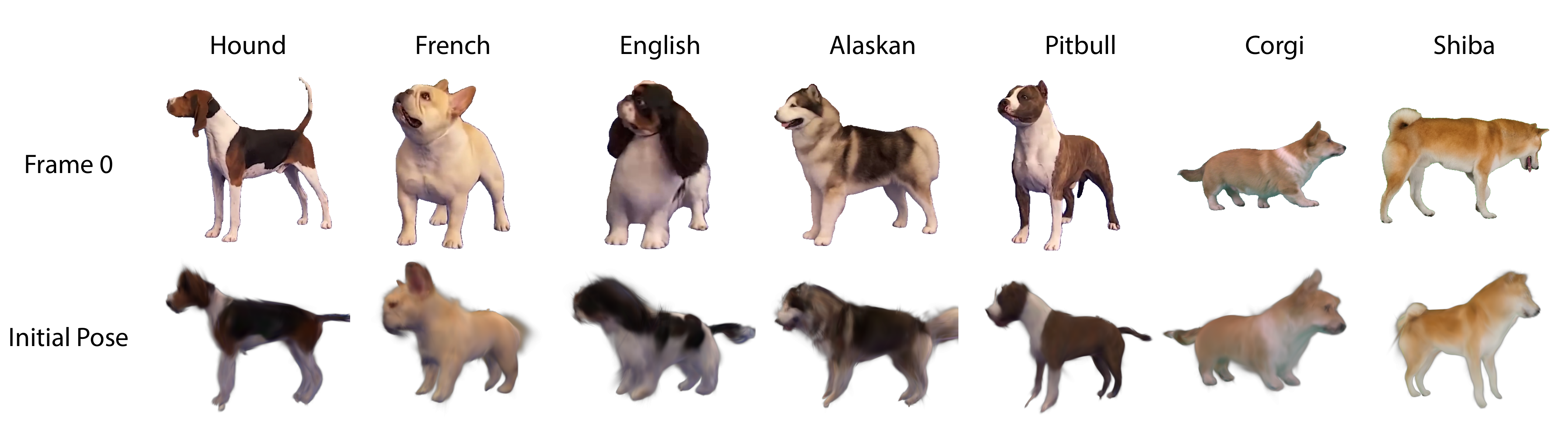}
    \caption{Top row: first supervision frame from the input videos. Bottom row: rendering of the reconstructed 3DGS asset in its GART rest pose, used as the initialization for our optimization. Notice the significant discrepancies in pose, outline, and color between the supervision and the initial model. These differences highlight the inherent complexity of this real-world setting.}
    \label{fig:GART_init_pose}
\end{figure}
Figure~\ref{fig:GART_init_pose} highlights the challenges in the {GART}~\cite{lei2024gart} setting. The input 3DGS model corresponds to the canonical asset pose reconstructed by GART, while the supervision frames depict the animal in motion, leading to geometric misalignment at initialization.
In addition, the model is reconstructed from in-the-wild videos captured under varying poses, viewpoints, zoom levels, and lighting conditions. Consequently, the reconstructed 3DGS model may exhibit color inconsistencies and appearance gaps relative to the supervision video.
Together, these factors create substantial geometric and photometric discrepancies between the initial model and the supervision frames, reflecting a realistic scenario without precise camera parameters or consistent illumination.

\subsection{{Additional Results}}
\begin{figure}[t]
    \centering
    \includegraphics[width=0.8\textwidth]{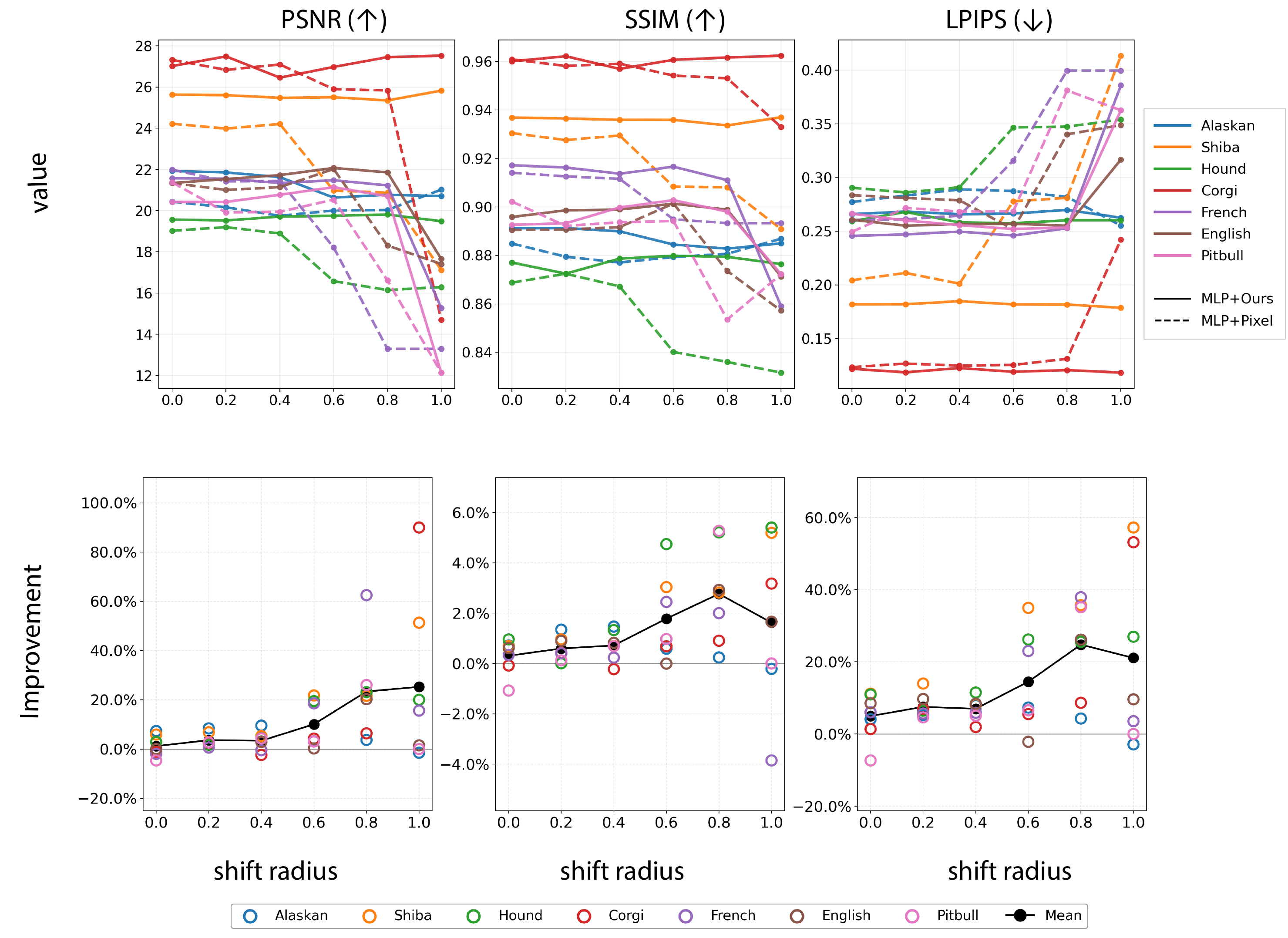}
    \caption{Robustness to initial spatial misalignment on the GART dataset.  \textbf{Top:} Per-dog metric curves for PSNR, SSIM, and LPIPS under increasing shift. The performance gap widens as the misalignment increases, demonstrating the stability and robustness of our method.
    \textbf{Bottom:} Mean performance gain in PSNR, SSIM, and LPIPS across all dogs as a function of the shift radius. The minimum and maximum values for each individual dog are shown as scattered points.}
    \label{fig:GART_graphs}
\end{figure}
We provide additional quantitative results of the GART misalignment experiment described in the main paper.
In Figure~\ref{fig:GART_graphs} (top), we present the per-dog plots for all metrics. The plot for each dog is shown in a different color, with dashed lines corresponding to pixel-only supervision (MLP+Pixel) and solid lines to our spectral scheme (MLP+Ours). Across nearly all dogs, pixel-based optimization degrades more rapidly with increasing shift, while the spectral method remains significantly more stable.

Figure~\ref{fig:GART_graphs} (bottom) compares the performance gap across dogs for the three metrics as a function of the shift radius. We report the mean improvement over all dogs, while the minimum and maximum values for each individual dog are shown as colored scatter points.
For PSNR and SSIM, we report the difference (MLP+Ours - MLP+Pixel), whereas for LPIPS we report (MLP+Pixel - MLP+Ours), so that positive values consistently indicate a performance gain of our method over MLP+Pixel. As the shift radius increases, the improvement steadily grows, highlighting the robustness of our approach under larger misalignment.

\paragraph{LPIPS Limitations.}
\begin{wrapfigure}{r}{0.5\textwidth}
    \centering
    \includegraphics[width=0.9\linewidth]{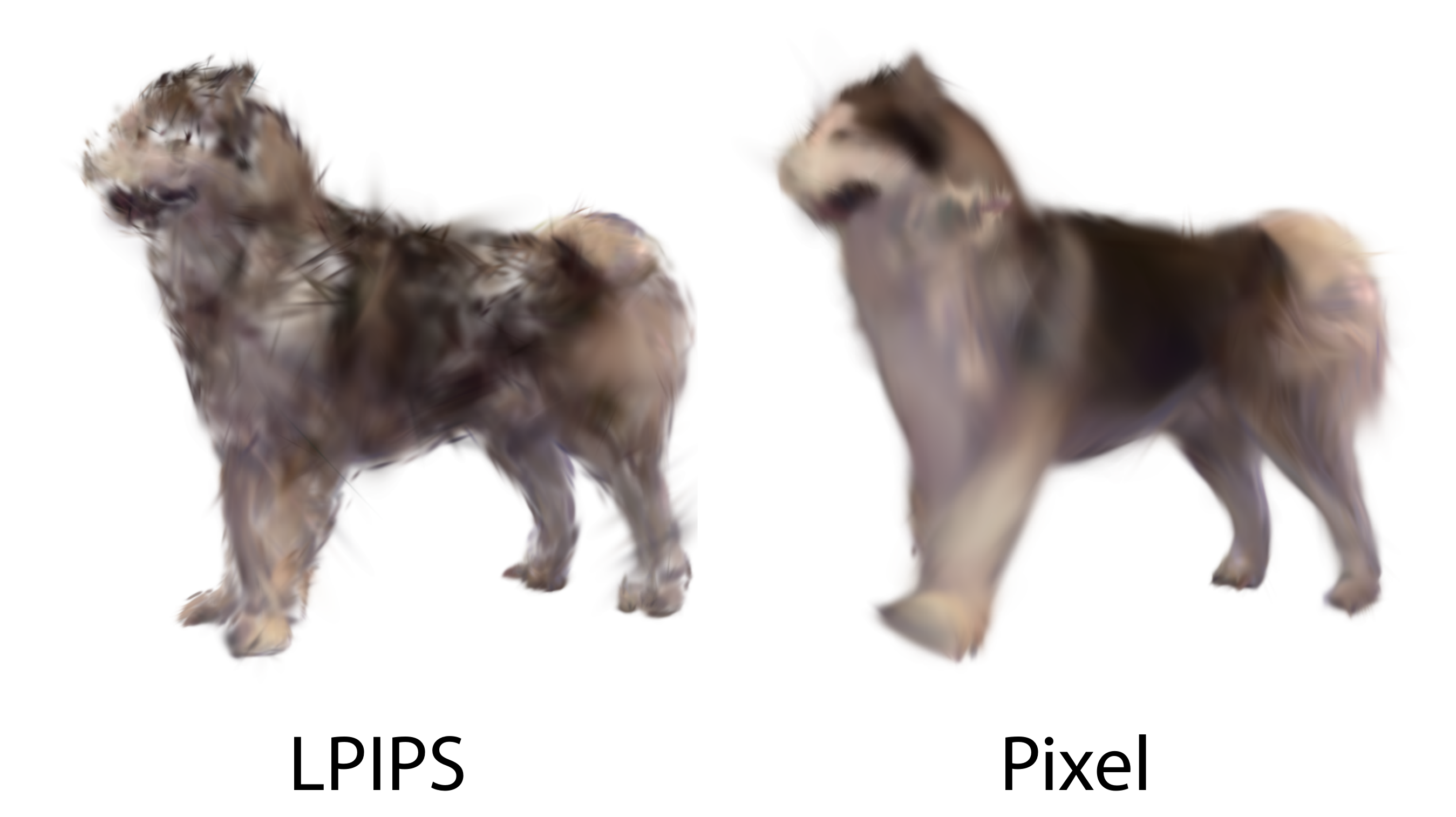}
    \caption{LPIPS vs.\ pixel supervision in the second phase on GART. Due to color discrepancies between the reconstructed 3DGS and the video, LPIPS provides weaker geometric constraints, resulting in blurrier results.}
    \label{fig:GART_warmup_inset}
\end{wrapfigure}
Following the SC4D experiment, we experimented with replacing the pixel-based loss with {LPIPS}~\cite{zhang2018perceptual} supervision 
during training. 
Figure~\ref{fig:GART_warmup_inset} illustrates a representative failure on the GART dataset.
Since LPIPS is calibrated to capture perceptual differences in color and luminance \cite{zhang2018perceptual}, it can interpret global color discrepancies as meaningful structural changes. In our setting, this leads to gradients that prioritize compensating for lighting gaps rather than enforcing geometric consistency, ultimately degrading the 3DGS optimization.

\clearpage
\section{Global Loss Baselines}

\subsection{Distance Transform Loss Formulation}
To adapt the distance transform proposed in \cite{liu2019soft} to our setting, we define it as follows.
Let $O_{rend} \in [0,1]^{H\times W}$ be the predicted soft mask (the alpha channel of the Gaussian rendering), and let $O_{gt} \in \{0,1\}^{H\times W}$ be the target ground-truth binary mask. The Euclidean distance transform of a binary image $B \in \{0,1\}^{H\times W}$, denoted by $DT(B)$, assigns to each pixel the Euclidean distance to its nearest zero-valued pixel. 
We compute two normalized distance maps for foreground (fg) and background (bg) regions:
$$D_{fg} = \frac{DT(1-O_{gt})}{\max_{i,j} DT(1-O_{gt})_{i,j}}, \qquad D_{bg} = \frac{DT(O_{gt})}{\max_{i,j} DT(O_{gt})_{i,j}},$$
and define the Distance Transform Loss as
$$L_{DT} = \frac{\left\| D_{fg} \cdot O_{rend} + D_{bg} \cdot (1 - O_{rend}) \right\|_1}{H\cdot W}$$
The first term produces gradients that push Gaussians away from background regions, while the second encourages dense coverage of the foreground.

\subsection{Qualitative Results}
\begin{figure}
\centering
\includegraphics[width=\linewidth]{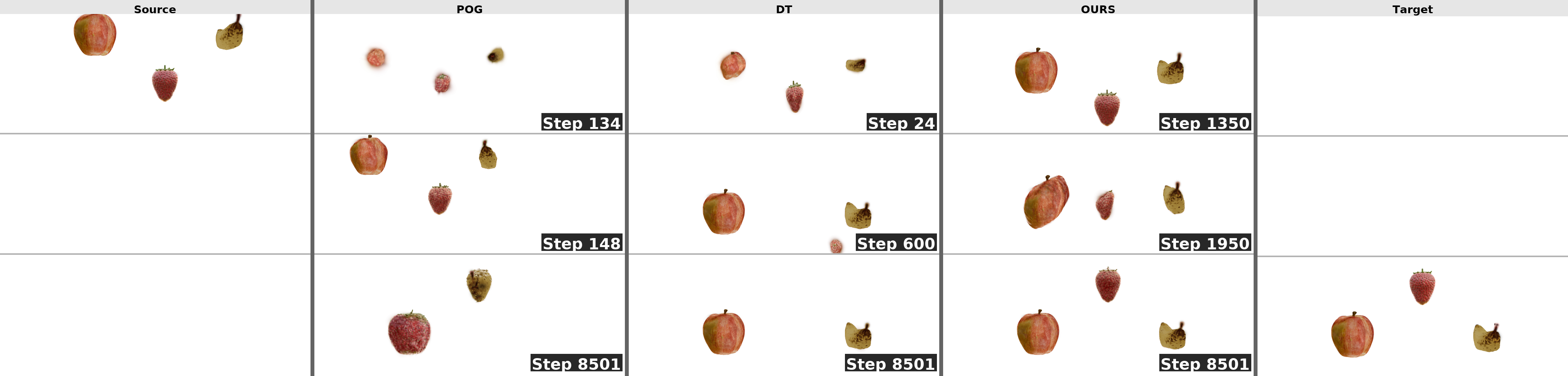}
\caption{Qualitative comparison against the global-loss baselines PoG and DT on the teaser scene. While both baselines exhibit object merging during optimization, our method converges correctly and preserves object separation.}
\label{fig:teaser_global_baselines}
\end{figure}
\noindent Beyond the quantitative comparison on GART, we provide a qualitative comparison against the global-loss alternatives Pyramid of Gaussians (PoG) and Euclidean Distance Transform (DT) on the teaser scene. Figure~\ref{fig:teaser_global_baselines} shows snapshots from the optimization process, illustrating the failure modes of both methods. In each case, the three objects gradually merge during optimization; for example, under DT, the center strawberry is absorbed into the banana model. In contrast, our method converges correctly while preserving object separation throughout optimization.

\clearpage
\section{Implementation Details}
\subsection{Hyperparameters.}
The main hyperparameters of our method govern the frequency annealing schedule and the transition from spectral to spatial supervision. A detailed description of these parameters is provided in Table~\ref{tab:param_meaning}, while their specific values for each experiment are reported in Table~\ref{tab:experiment_params}.
Across all experiments, we use 800 control points and train the model for 10K iterations. We fix the global image loss weight \texttt{lambda\_image} ($\lambda_{\text{image}}$) to 5000 and set \texttt{arap\_start\_iter} to 1000.

\begin{table}[h]
\caption{Description of training hyperparameters and their roles.}
\centering
\small
\renewcommand{\arraystretch}{1.15}
\setlength{\tabcolsep}{3pt}
\begin{tabular}{l p{8cm}}
\hline
\textbf{Parameter} & \textbf{Meaning} \\
\hline

\texttt{add\_pixel\_loss} 
& Iteration at which training switches from spectral loss to spatial (pixel) loss. \\

\texttt{num\_bands} 
& Number of frequency bands used in the spectral basis. \\

\texttt{warmup} 
& Percentage of training iterations during which only the first frequency band is active. \\

\texttt{arap\_start\_iter} 
& Iteration at which ARAP regularization is introduced. \\

\texttt{lambda\_arap} ($\lambda_{\text{arap}}$) 
& Weight controlling the strength of the ARAP regularization term. \\

\texttt{lambda\_spec\_mask} ($\lambda_{s_{\text{mask}}}$) 
& Relative weight of the spectral mask loss term. \\

\texttt{lambda\_bce} ($\lambda_{\text{bce}}$) 
& Weight of the binary cross-entropy term in $\mathcal{L}_{\text{image}}^{\text{pixel}}$. \\

\texttt{lambda\_late\_lpips} 
& Weight of the LPIPS term when it replaces the spatial $L_2$ loss. \\

\texttt{deform\_lr\_init} 
& Initial learning rate for the deformation learning. \\

\texttt{deform\_lr\_final} 
& Final learning rate for the deformation learning. \\

\hline
\end{tabular}
\label{tab:param_meaning}
\end{table}

\begin{table}[h]
\caption{Hyperparameter settings for SC4D (MLP), SC4D (Direct Morph Field), and GART experiments.}
\centering
\small
\setlength{\tabcolsep}{4pt} 
\renewcommand{\arraystretch}{1.1} 
\begin{tabular}{l c c c}
\hline
\textbf{Parameter} & 
\textbf{SC4D (MLP)} & 
\textbf{SC4D (Direct Morph)} & 
\textbf{GART} \\
\hline

add\_pixel\_loss   & 6000   & 4000   & 7000 \\
lambda\_spec\_mask & 0.5    & 0.2    & 0.3  \\
lambda\_bce        & 0.3    & 0.3    & 0.1  \\
warmup             & 0.3    & 0.2    & 0.25 \\
lambda\_arap       & 1.0    & 3.0    & 1    \\
deform\_lr\_init   & 0.0002 & 0.001  & 0.001 \\
deform\_lr\_final  & 0.0001 & 0.0005 & 0.0005 \\
num\_bands         & 8      & 6      & 8 \\
\hline
\end{tabular}
\label{tab:experiment_params}
\end{table}

\subsection{Runtime \& Other Details.} 
All experiments were conducted on a single NVIDIA L40 GPU. Training a single sequence requires approximately 8–15 minutes, depending on the dataset and configuration. All other technical implementation details follow the original GSGD setup.
We note that our method adds negligible overhead over pixel-based loss (e.g., 437s vs. 443s on SC4D), remaining within 0-6\% across experiments.


\end{document}